\documentclass[11pt,a4paper]{article}
\usepackage{times,latexsym}
\usepackage{url}
\usepackage[T1]{fontenc}

\usepackage[acceptedWithA]{tacl2021v1}

\usepackage{xspace,mfirstuc,tabulary}

\newif\iftaclinstructions
\taclinstructionsfalse 
\iftaclinstructions
\renewcommand{\confidential}{}
\renewcommand{\anonsubtext}{(No author info supplied here, for consistency with
TACL-submission anonymization requirements)}
\newcommand{\instr}
\fi

\iftaclpubformat 

\else

\fi

\usepackage{amsmath}
\usepackage{amssymb}
\usepackage{booktabs}
\usepackage{graphicx}
\usepackage{multirow}
\usepackage{multicol}
\usepackage{cleveref}
\usepackage{makecell}
\usepackage{tabularx}
\usepackage{longtable}
\usepackage{xtab}
\usepackage{seqsplit}
\usepackage[utf8]{inputenc}
\usepackage{array}
\usepackage{colortbl}
\usepackage{xcolor}
\usepackage{hhline}
\usepackage{microtype}
\usepackage{subcaption}
\usepackage{tcolorbox}
\tcbuselibrary{listings,breakable}
\usepackage{enumitem}
\usepackage[T1]{fontenc}
\usepackage{textcomp}
\usepackage{upquote}
\usepackage{float}
\usepackage{listings}
\definecolor{darkblue}{rgb}{0, 0, 0.5}
\hypersetup{colorlinks=true, citecolor=darkblue, linkcolor=darkblue, urlcolor=darkblue}

\newcommand{\bench}[0]{{\textsc{PhantomBench}}}
\newcommand{\halluc}[1]{\textcolor{red!75!black}{#1}}
\newcommand{\correct}[1]{\textcolor{blue!70!black}{#1}}

\newtcolorbox{prompttext}[1]{
    arc=2mm, boxrule=0.5pt,    
    colback=gray!5, colframe=gray!50, 
    title={#1}, fonttitle=\bfseries\small,
    fontupper=\small,
    top=2pt, bottom=3pt, left=2pt, right=4pt,
    toptitle=2pt, bottomtitle=2pt,     
    before upper=\noindent,
}
\newtcblisting{promptcode}[1]{
    arc=2mm, boxrule=0.5pt,    
    colback=gray!5, colframe=gray!50, 
    title={#1}, fonttitle=\bfseries\small,
    listing only,             
    top=2pt, bottom=2pt, left=2pt, right=4pt,
    toptitle=2pt, bottomtitle=2pt,     
    listing options={
        basicstyle=\footnotesize\ttfamily,
        breaklines=true,      
        breakindent=0pt,      
        columns=flexible,
    },
}

\crefformat{section}{\S#2#1#3}
\Crefformat{section}{\S#2#1#3}

\crefformat{subsection}{\S#2#1#3}
\Crefformat{subsection}{\S#2#1#3}

\title{\bench{}:\\
Benchmarking the Non-existential Threat of Language Models}

\author{
  Haeji Jung$^1$ 
  \and
  Hila Gonen$^{1,2}$
  \\
  \ \\
  $^1$University of British Columbia
  \\
  $^2$Canada CIFAR AI Chair, Amii
  \\
  \texttt{\{haejij, hgonen\}@cs.ubc.ca}
}

\date{}

\begin{document}
\maketitle
\begin{abstract}

Hallucinations, where language models (LMs) generate factually ungrounded responses, pose serious risks, as users tend to blindly rely on them. This is particularly concerning in high-stakes domains, where consequences of such model behavior can lead to significant harms. Despite notable progress in understanding hallucinations, it remains unclear how reliably these models can recognize the limits of their knowledge.
We introduce \bench{}, the first large-scale benchmark of its kind, comprising more than 60K non-existent terms and entities derived from real concepts across diverse domains. 
Using our benchmark, we evaluate a total of 23 models of various types and sizes.
We show staggering hallucination rates across the board (with average rates as high as 86.7\% in some cases), and note that even frontier models surprisingly fail to abstain on non-existent concepts, especially when the input presumes their existence.
We then show that \bench{} can serve as a proxy for studying model behavior on rare concepts for which models are more prone to hallucinate.
We also provide a pipeline to construct \bench{}, enabling scalable generation of non-existent concepts tailored to the specific needs of researchers and practitioners.\footnote{The pipeline and benchmark will be released upon publication.}

\end{abstract}
\section{Introduction}

\begin{figure*}[t]
    \centering
    \includegraphics[width=\textwidth]{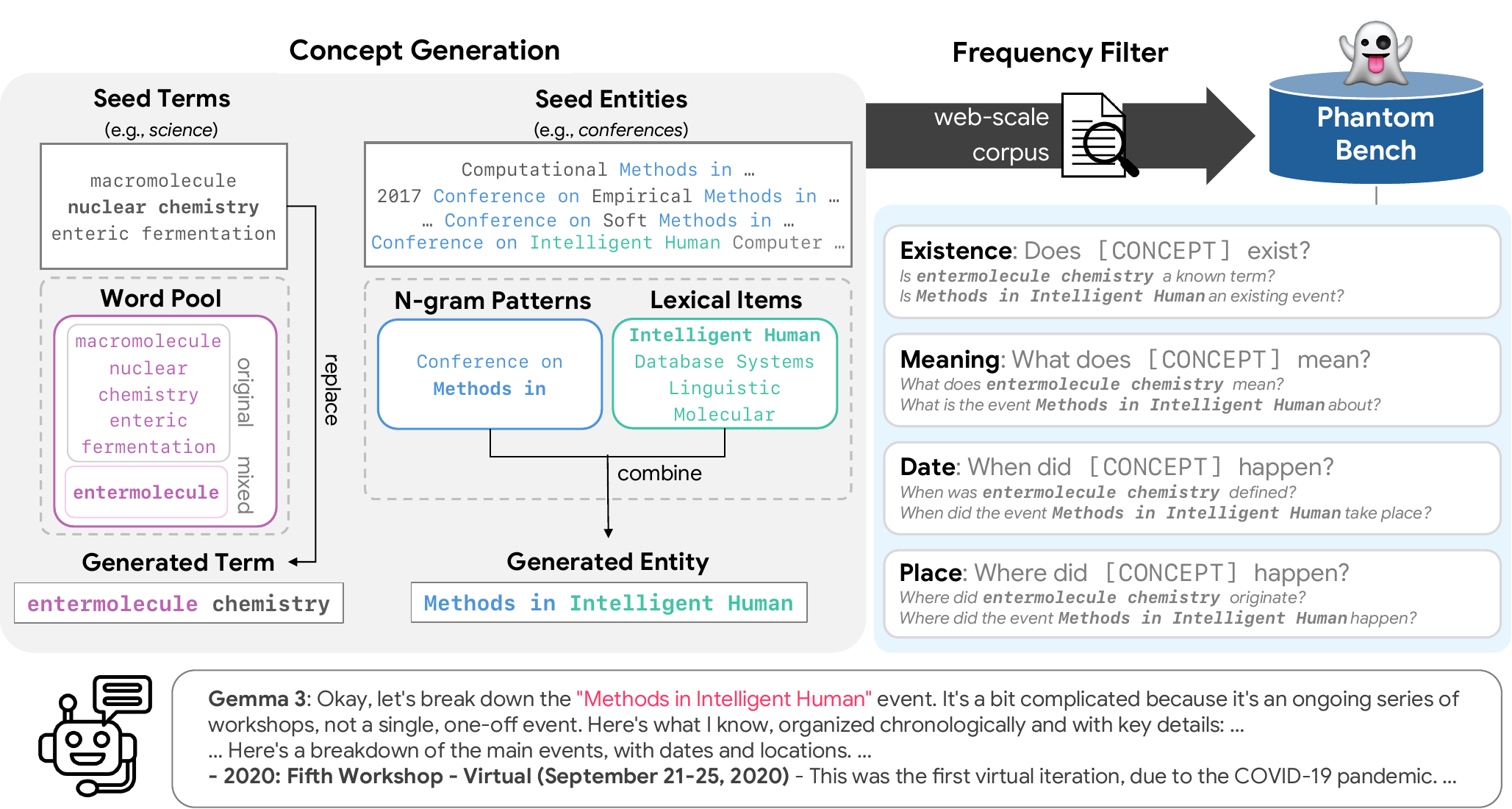} 
    \caption{The pipeline to construct \bench{}. Existing concepts from seed terms and entities are decomposed into smaller components (words and $n$-grams) which are then recombined to form new concepts (\Cref{subsec:pipeline}). Frequency filter discards concepts found in a large corpus, considering concepts with zero matches as \textit{non-existent} (\Cref{subsec:nonexistence_verification}).
    The resulting concepts are queried through diverse prompts targeting different attributes of the concept (\Cref{subsec:prompt_level}). At the bottom is an example response from Gemma~3-12B to a generated non-existent entity, \textit{Methods in Intelligent Human}.
    }
    \label{fig:concept_generation}
\end{figure*}

Language models (LMs) have been shown to generate responses that are not grounded in factual information, despite substantial advances in their capabilities. 
This phenomenon, commonly referred to as hallucination, remains a persistent challenge for building reliable LM systems~\cite{huang-etal-2023-survey,liu-etal-2025-unified,kalai-etal-2025-why}.
In particular, recent work demonstrates that language models tend to generate plausible-sounding answers when queried with inputs that fall outside their knowledge boundary \cite{li-etal-2025-knowledge-boundary,wen-etal-2025-know}.
This becomes particularly critical as LMs are increasingly deployed in real-world settings, where input distributions are inherently noisy and often include ill-defined, ambiguous, or even non-existent concepts.
The risk is even greater in high-stakes domains such as healthcare and law, and is further amplified when models are deployed at a massive scale, where generating plausible but ungrounded information can lead to serious consequences \cite{ho-legal,risk}.
Despite its critical importance, the ability of LMs to recognize and appropriately abstain from answering unanswerable queries remains an open challenge \cite{brahman-etal-2024-art,kirichenko-etal-2026-abstentionbench}.
Evaluating such behavior is further complicated by the difficulty of reliably identifying hallucinations in free-form model outputs \cite{min-etal-2023-factscore,bang-etal-2025-hallulens}.

In this paper, we introduce \textbf{\textsc{PhantomBench}}, the first large-scale benchmark of its kind, designed to evaluate LM abstention using plausible but non-existent concepts that are derived from existing ones across multiple domains. 
Our benchmark enables a straightforward and reliable hallucination evaluation scheme: providing any information about a non-existent concept is hallucination by definition.
To support diverse evaluation settings, we curate structured subsets that vary along key dimensions such as concept type and domain, enabling focused analyses of specific model behaviors. 
Importantly, we propose a scalable data generation pipeline that can be adapted to new domains or seed concepts, accompanied by a human validation of the generated concepts. This enables researchers and practitioners to construct customized benchmarks tailored to their specific use cases, and also ensures the benchmark remains applicable over time, without relying on a fixed set of concepts that might eventually appear in future training data.

We evaluate six widely used language models on the full benchmark and find that all of them struggle to reliably abstain from answering queries about non-existent concepts, as illustrated by the example response in \Cref{fig:concept_generation}. Further evaluation across different model types and sizes using various dedicated subsets of the benchmark shows that even larger models, reasoning-based models and domain-specialized models, often fail to abstain more frequently than smaller general-purpose ones.
Finally, we compare the abstention behavior of models on non-existent and existing concepts, and conclude that non-existent concepts can serve as a proxy for existing rare concepts. This is particularly important because systematically evaluating models on unfamiliar or low-frequency knowledge remains challenging, despite such scenarios often being sensitive and especially prone to hallucination.

Our contributions are summarized as follows: (a) We introduce a large-scale benchmark -- \bench{}, consisting of over 60K non-existent concepts to evaluate language model abstention (\Cref{sec:phantombench}); (b) We provide a scalable concept generation pipeline that can be applied to any set of seed concepts in any domain (\Cref{sec:phantombench}); (c) We evaluate 23 models across different model families, sizes, reasoning capabilities, and domain specialization, showing that even models expected to be more reliable, struggle to abstain appropriately (\Cref{sec:results}); (d) We show that non-existent concepts can serve as a practical proxy for studying model behavior on rare concepts (\Cref{sec:selective_abstention}).
Beyond evaluating abstention on non-existent concepts, \bench{} reveals systematic patterns of model behavior and provides a scalable testbed for studying reliability on rare and unknown concepts.

\section{\textsc{PhantomBench}}\label{sec:phantombench}

We build \textsc{PhantomBench} with non-existent concepts, both \textit{terms} and \textit{entities}, to evaluate whether language models abstain when presented with input beyond their knowledge. We distinguish between terms and entities: terms refer to abstract concepts typically used in domain-specific contexts (e.g., \textit{nuclear chemistry}), whereas entities refer to specific identifiers that point to a unique, existing object or event (e.g., \textit{Computational Methods in Systems Biology}), allowing us to employ generation strategies tailored to each concept type. 

\paragraph{Pipeline Overview} We design a pipeline that constructs non-existent yet linguistically plausible concepts, while enabling controllable and extensible generation.
Specifically, we first generate candidate concepts by combining words or word fragments from existing concepts to ensure a linguistically plausible structure (\Cref{subsec:pipeline}). We then filter out any instances that appear in a web-scale corpus to obtain the final set of non-existent entries (\Cref{subsec:nonexistence_verification}). Lastly, we form prompts with varying difficulty levels to query the model (\Cref{subsec:prompt_level}).
Our extensible pipeline enables new concepts to be easily generated for any target domain or set of seed concepts, allowing the benchmark to remain relevant over time, and for targeted usages. We validate the pipeline through a human study of the generated concepts in \Cref{subsec:benchmark_construction}.

\subsection{Non-existent Concept Generation}
\label{subsec:pipeline}

We apply different strategies to create non-existent concepts depending on their types (i.e., \textit{terms} and \textit{entities}). \Cref{fig:concept_generation} shows the framework for generating concepts of each type, as described below.

\paragraph{Term Generation} 
To generate new terms, we first extract all words from a set of existing terms $\mathcal{T}_e$ in the source data to form a set of words $\mathcal{W}_e$.\footnote{We define a word as a unit separated by white spaces.} 
Then we construct a set of blended words $\mathcal{W}_g$ by combining parts from different existing words. For example, the word \textit{entermolecule} is generated by combining two existing words: \textit{enteric} and \textit{macromolecule}. We maximize plausibility by splitting words in a way that preserves common affixes (see \Cref{appx:term_mix_details} in the Appendix for details). As this process can lead to a quadratic number of combinations with respect to the number of source words, we limit the number of blended words with a hyperparameter. Using the full set of resulting words $\mathcal{W}=\mathcal{W}_e \cup \mathcal{W}_g$ (existent and generated), we replace a subset of words in an existing term $t \in \mathcal{T}_e$ to create a new non-existent term. For example, we replace \textit{nuclear} with \textit{entermolecule}, to generate the non-existent term \textit{entermolecule chemistry} based on the original term \textit{nuclear chemistry}.
We replace half of the words in $t$ with the same number of words sampled from $\mathcal{W}$, so that longer terms have more words substituted, increasing lexical variety while maintaining term plausibility.

\paragraph{Entity Generation} 
Entity names are often compositional, consisting of a recurring structural pattern and a semantically specific lexical item (e.g., \textit{Geneva} (semantically specific) + \textit{International Music Competition} (structural)).
We generate new entities by combining these two components extracted from existing entities. Inspired by prior observations that term frequency correlates positively with structural productivity \cite{bybee} and inversely with semantic specificity \cite{zipf,zipf-review}, we treat high-frequency $n$-gram patterns as structural patterns and lower-frequency ones as lexical items.
To construct a list of structural patterns, we set a threshold $t_{ngram}$ and store bigrams and trigrams with higher frequencies than this threshold for each category (e.g., ``Methods in'' under \textit{conferences} category and ``Battle for'' under \textit{historical events}).
For lexical items, we select $n$-grams whose frequencies fall between a lower threshold and $t_{ngram}$ from the same category. In order to keep lexical items suitable as semantic cores, we apply additional constraints, such as filtering out those with predefined stopwords (e.g., \textit{of}, \textit{the}, \textit{and}).
Examples of extracted lexical items include ``Intelligent Human'' and ``Wagner''.
We then combine sampled lexical items with the structural patterns to form new entities such as \textit{Methods in Intelligent Human} and \textit{Battle for Wagner}.
Further details on $n$-gram extraction and lexical item constraints are provided in \Cref{appx:entity_mix_details}. 

\subsection{Non-existence Verification}\label{subsec:nonexistence_verification} 
After generating the concepts, we filter out concepts that are likely to exist. We operationalize \textit{non-existence} as \textit{unattested}: a concept not found in a large corpus is labeled as non-existent, while those with exact matches are considered existing and is filtered out.
Since existing domain-specific terms and named entities tend to be documented somewhere, the absence of even a single occurrence in a corpus of billions of documents is a reasonable proxy for non-existence. The search is case-sensitive and covers several casing variants of each concept.
More details and analyses of this stage are provided in \Cref{subsubsec:frequency_estimation} and \Cref{appx:freq_distribution}.

\subsection{Prompt Design}\label{subsec:prompt_level}
\begin{table}[]
    \centering
    \resizebox{\columnwidth}{!}{%
    \setlength{\tabcolsep}{3pt}
    \begin{tabular}{llp{6.4cm}}
    \toprule
        Property & Type & Examples\\\midrule
       \multirow{2}{*}{Existence} & Term & Does the term `X' actually exist?\\
       &Entity&Is there anything called `X'?\\\midrule
       \multirow{2}{*}{Meaning} & Term & What does `X' mean?\\
       &Entity& What does `X' refer to?\\\midrule
       \multirow{2}{*}{Date} & Term & When did the term `X' originate?\\
       &Entity& When was `X' first introduced?\\\midrule
       \multirow{2}{*}{Place} & Term & Where was `X' discovered?\\
       &Entity&Where did `X' take place?\\\midrule
       Etymology$^{*}$ & Term & Why was the name `X' chosen for this concept?\\\midrule
       Application$^{*}$ & Term & What are the primary advantages of `X'?\\\midrule
       Relation$^{*}$ & Term & What are the three most similar entities to `X'?\\\bottomrule
    \end{tabular}%
    }
    \caption{Prompt examples used to query language models. X denotes the queried concept, and $^{*}$ denotes additional attribute types applied only to subsets of terms.}
    \label{tab:prompt_examples}
\end{table}
We design prompts that either explicitly query the existence of a concept or implicitly assume its existence by asking about one of its attributes. This allows us to analyze how models respond to questions about non-existent concepts under different forms of presupposition. For example, ``Does \texttt{[CONCEPT]} exist?'' does not assume the existence of the concept, whereas queries such as ``Where did \texttt{[CONCEPT]} happen?'' or ``When was \texttt{[CONCEPT]} established?'' implicitly presume that it exists.

We predefine four properties applicable to both terms and entities: existence, meaning, date, and place. 
To further analyze the impact of different queried attributes, we introduce additional attributes for terms: etymology, application, and term relation.
The full list of queried properties is provided in \Cref{tab:prompt_examples}, along with example prompts.
\section{Evaluation Protocol}\label{sec:evaluation}

\paragraph{LLM-as-a-Judge} 
In our setting, we deal with open-ended responses, reflecting real-world usage in which users expect unconstrained generations. However, automatic evaluation of such responses is challenging due to the diversity and ambiguity of possible abstaining behaviors.
We therefore employ LLM as a judge to enable scalable evaluation of open-ended generations.
Specifically, we require the judge model to perform a binary decision on whether the model response is abstaining from answering the question. 
The prompt we use for the LLM judge is provided in \Cref{appx:judge_prompt} in the Appendix.

We validate the quality of the LLM judge through a statistical study on a sample of model responses. We sample 120 prompt-response pairs on non-existent concepts across datasets and prompt types, and recruit four human annotators\footnote{The annotation was conducted by four graduate students, including one author and three volunteer lab members with NLP research experience.} to determine whether each response abstains or not. We then conduct Alternative Annotator Test proposed by \citet{alt-judge} to validate the alignment between judgments of the LLM and human annotators. The LLM judge achieved a winning rate of 1.00 (100\%), demonstrating that it agreed with the remaining human annotators just as often as (if not more than) any single human did. Detailed setup and results are provided in  \Cref{appx:justify_llm_judge} in the Appendix.

\paragraph{Evaluation Metric}
We quantify models' performance using hallucination rate (HR), defined as the proportion of queries for which the model produces a non-abstention response. Formally,
\begin{equation}
    \text{HR} = \frac{1}{N} \sum_{i=1}^{N} \mathbf{1}[\text{Abstain}(r_i)=\texttt{False}]
\end{equation}
where $N$ is the number of instances and $r_i$ is the response for $i$-th instance.

\section{Experimental Setup}
\label{sec:experiments}

\begin{table*}[t]
    \centering
    \resizebox{0.9\textwidth}{!}{%
    \begin{tabular}{@{}llrl@{}}
    \toprule
        Source & Category  & \makecell[r]{\# Generated \\Concepts} & Generated Examples \\\midrule
        \multicolumn{4}{l}{\textit{\textbf{Terms}}}\\\addlinespace[2pt]\midrule
         \multirow{3}{*}{\makecell[l]{MedINST \\\citep{han-etal-2024-medinst}}} & MeDAL &28,347 & urodynamic carotid stenosis, levamisole dermatitis \\\cmidrule{2-4}
         & NCBI-disease & 1,626&hemophilia lymphoma, myelopathy carcinomas \\\cmidrule{2-4}
         & UMNSRS & 302 & Airsicol, Clutterine, Convulorrhea \\\midrule
         Wikipedia & \makecell[l]{Glossaries of Science} & 5,901& vibrator third palatalization, pericline circuit \\\midrule
         Wiktionary &English legal terms & 725& contempt bonis, affirmoxy, reversion nullius \\\midrule
        \multicolumn{4}{l}{\textbf{\textit{Entities}}}\\\addlinespace[2pt]\midrule
        \multirow{11}{*}{\makecell[l]{Wikidata \\Event Instances}} & Festival & 920 & Melbourne Screams Short Film Festival \\\cmidrule{2-4}
        & Conference & 4,455 & Conference on Technology Asia-Pacific Digital  \\\cmidrule{2-4}
        & Holiday & 340 & Octave of Florian  \\\cmidrule{2-4}
        & Sport Event & 4,488 & triathlon at the Shooting Championships  \\\cmidrule{2-4}
        & Competition & 3,981 & Bundesvision Song Contest Twin Peaks  \\\cmidrule{2-4}
        & Show / Exhibition & 3,316 & London Runway International Auto Fashion Show \\\cmidrule{2-4}
        & Election & 3,567 & Polish Amarante municipal election \\\cmidrule{2-4}
        & Social Issue & 1,854 & Dock Hill miners' strike \\\cmidrule{2-4}
        & Natural Disaster & 300 & Ava Tropical depression \\\cmidrule{2-4}
        & Accident & 680 & Delta Air train crash\\\cmidrule{2-4}
        & Historical Event & 1,168&Hama Battle of Fort\\\midrule
         \makecell[l]{Pop QA \\\citep{popqa}} & \makecell[l]{Creative Work / Place \\(\texttt{PER} removed)} & 441 & Rock The Great Crime, You Are My Heart Places\\
         \bottomrule
    \end{tabular}%
    }
    \caption{Sources and statistics of generated terms and entities, along with examples.}
    \label{tab:original_datasets}
\end{table*}

\subsection{Benchmark Construction}\label{subsec:benchmark_construction}
\paragraph{Source Datasets} 
We employ 17 datasets as sources of seed concepts. \Cref{tab:original_datasets} lists the datasets used to construct our benchmark, along with the statistics of the generated concepts. PopQA-non-human, MeDAL, NCBI-disease, and UMNSRS are sourced from publicly available datasets, while the remaining sources (Event Instances, Glossaries of Science, and English legal terms) are derived from Wikimedia sources.
The datasets used for term generation include domain-specific datasets from the medical, scientific, and legal domains. The datasets used for entity generation include names of events, creative works (e.g., books, songs, etc.), and places. Details of the data collection process are provided in \Cref{appx:data_details} in the Appendix.

\paragraph{Frequency Estimation}\label{subsubsec:frequency_estimation}
As described in \Cref{subsec:nonexistence_verification}, we verify non-existence of a generated concept based on web-scale corpus search, where we consider a concept to be non-existent if it has zero exact matches in the corpus. 
We use Dolma v1.7 \citep{soldaini-etal-2024-dolma} as the reference database, which contains more than 2.6 trillion tokens across 3.4 billion documents spanning 15 sources, including a mix of web content, academic publications, code, books, and encyclopedic materials. For efficient large-scale search, we use Infini-gram \citep{infinigram}, which supports fast $n$-gram search over massive corpora. For each generated entry, we perform an exact-match search and discard any entry that appears in the corpus. Since Infini-gram only supports case-sensitive search, we sum the number of matches across four casing variations to approximate case-insensitive search: \texttt{original} (as generated), \texttt{Title} (capitalize every word), \texttt{Capitalize} (capitalize only the first letter), and \texttt{lower}. The number of filtered entries and the distribution of their frequencies can be found in Appendix \Cref{appx:freq_distribution}.
As a robustness check, we run two additional non-existence checks on a subset of concepts (see Appendix \Cref{appx:gemini_check} and \Cref{appx:variant_check}).

\paragraph{Benchmark Targeted Splits}
\textsc{PhantomBench} consists of 62,411 non-existent concepts including 36,901 terms and 25,510 entities.
To support a range of analyses, we derive several targeted subsets from the full benchmark. Specifically, \textbf{\textsc{Phantom-T}} and \textbf{\textsc{Phantom-E}} contain terms and entities, respectively, each covering concepts across multiple categories. \textbf{\textsc{Phantom-Med}} and \textbf{\textsc{Phantom-Legal}} are subsets consisting of terms relevant to the medical and legal domains, and are curated to evaluate domain-specialized models in their respective domains. Each subset contains approximately 1,000 concepts. Detailed statistics of each subset are provided in \Cref{appx:subsets} in the Appendix.

\paragraph{Suitability of Generated Concepts} \label{subsec:quality_verification_survey}

After generating non-existent concepts, we recruited two fluent English speakers to evaluate the plausibility and specificity of generated concepts on a 5-point scale. Plausibility measures whether concepts appear realistic and well-formed, while specificity measures whether they are semantically specific rather than overly generic (e.g., \textit{purple vegetable}). 
We compared ratings for generated and rare existing concepts using the Mann--Whitney U test. For terms, we found no statistically significant differences in plausibility or specificity. For entities, plausibility was comparable, but generated entities were rated significantly lower in specificity. This suggests that while our pipeline produces plausible entities, capturing the granularity of real-world entities remains challenging. We leave further refinement of entity specificity to future work. Full instructions, inter-annotator agreement, and results are provided in \Cref{appx:pipeline_validation} in the Appendix.

\subsection{Model Evaluation}\label{subsec:models}

In this section, we introduce the models we evaluate on \bench{}. Given the scale of the benchmark, we select a set of core models for evaluation on the full benchmark and evaluate additional models on selected subsets for targeted analyses. 
In all models, we use their instruction-tuned variants.

\paragraph{Core Models}
We employ six different models for comprehensive evaluation on our full benchmark: Llama~3.1-8B \citep{llama3}, Gemma~2-9B \citep{gemma2}, Gemma~3-12B \citep{gemma3}, Qwen~2.5-7B \citep{qwen25}, Qwen~3-8B \citep{qwen3}, and Mistral~7B-v0.3 \citep{mistral7b}.
We select the models most widely used within the model family,\footnote{Models with the most downloads within their model family as of Apr 2026 on Hugging Face (\href{https://huggingface.co/}{https://huggingface.co/}).} owing to their balance between efficiency and capability. 

\paragraph{Analysis Models} For more focused analyses, we employ several other models to evaluate on subsets of \bench{}.
In order to analyze how different model sizes lead to different abstention behavior, we employ Qwen 3 family (1.7B, 4B, 8B, 14B, and 32B) and Llama 3 (3.1-8B and 3.3-70B). We also evaluate proprietary Gemini models: Gemini 2.5 Flash and Pro \citep{gemini25}, Gemini 3.5 Flash \citep{gemini35flash}, and Gemini 3.6 Flash \citep{gemini36flash}. We additionally include OLMo-7B~\cite{groeneveld-etal-2024-olmo}, which is pre-trained on Dolma v1.7, the corpus used for frequency estimation during the construction of \bench{}.
To investigate reasoning models, we use DeepSeek-R1-Distill-Qwen-32B \citep{deepseekr1} and GPT-OSS-20B \citep{gptoss} with varying reasoning levels. We also evaluate domain-specialized models to see if they are better able to abstain towards non-existent terms within the domain (i.e., created based on in-domain seed terms). To this end, we employ BioMistral~7B \citep{biomistral} and MedGemma~4B \cite{medgemma} for the biomedical domain, and SaulLM-7B \citep{saullm7b} for the legal domain, along with their base counterparts, Gemma~3-4B \cite{gemma3} and Mistral~7B-v0.1 \cite{mistral7b}. 

\paragraph{Judge Model}
For the LLM judge, we use Gemini 2.5 Flash due to its strong performance and reliability for response evaluation. Since \bench{} contains more than 60K concepts whose abstention behavior must be evaluated entirely by the judge model, scalability is an important consideration. Among the Gemini 2.5 models, Flash provides lower latency and cost while still meeting our evaluation requirements. 

\begin{figure}[t]
    \centering
    \includegraphics[width=0.98\linewidth]{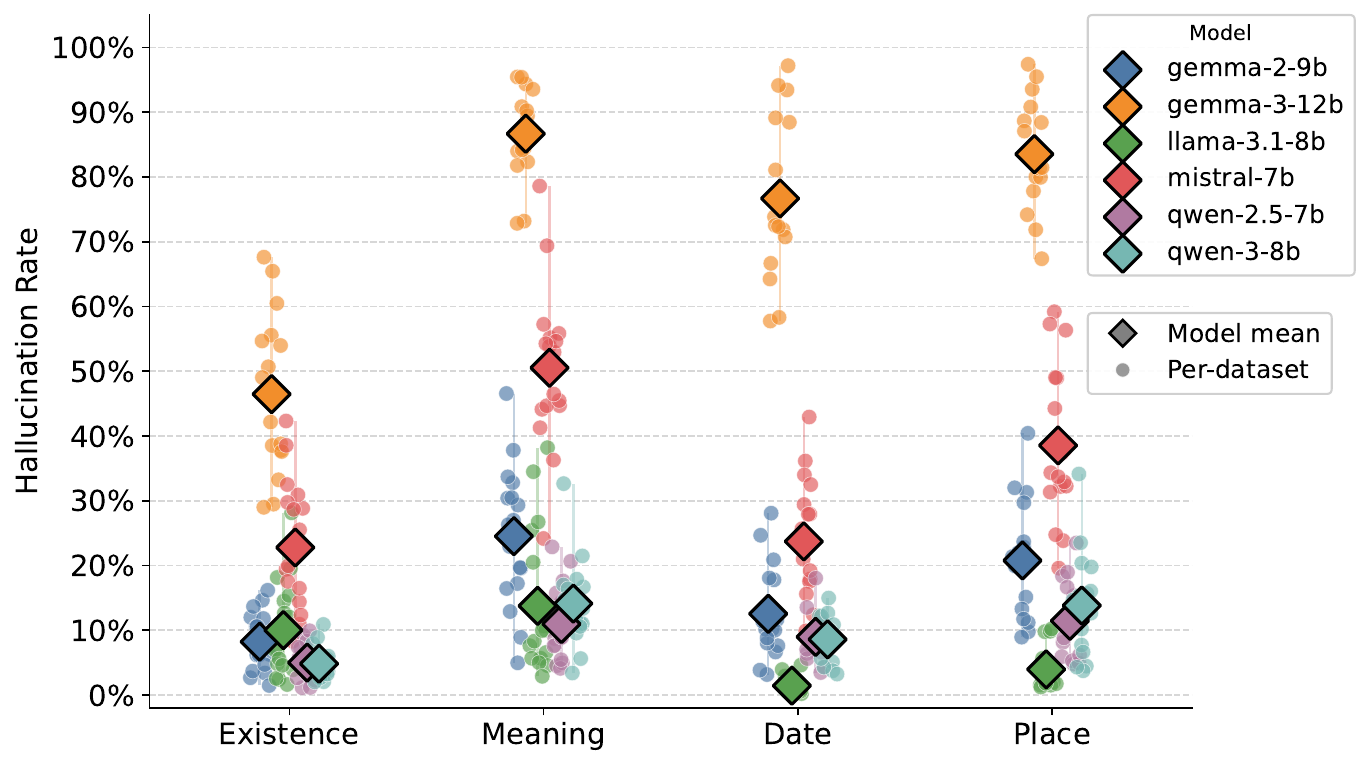}
    \caption{Hallucination rates by prompt type on non-existent terms and entities.}
    \label{fig:sota_results}
\end{figure}

\section{Evaluation Results}\label{sec:results}
In what follows, we present the results on the full benchmark (\Cref{subsec:overall_result}), and then turn to targeted analyses on selected subsets (\Cref{subsec:prompt_result}--\Cref{subsec:domain_result}) and the reliability of abstaining responses (\Cref{subsec:qualitative_analysis}). 

\subsection{All Models Fail to Reliably Abstain when Queried about Non-Existent Concepts}\label{subsec:overall_result}

\Cref{fig:sota_results} shows the results of core models on the full benchmark. 
All models struggle to abstain, especially when queried about the \textit{meaning} of a concept (HR of 33.4\%), even though they often correctly acknowledge that the concepts do not exist when queried explicitly about their \textit{existence} (HR of 16.2\%). We investigate the impact of prompted attribute in \Cref{subsec:prompt_result}.

Among six core models, Gemma 3 and Mistral show the highest hallucination rates at 73.3\% and 33.9\%, respectively. In contrast, Llama 3.1 8B and Qwen 2.5 are the most reliable at abstaining, with hallucination rates of 7.3\% and 9.1\%.
We further investigate how selectively these models abstain on non-existent concepts, and find that models with high abstention rates on non-existent concepts also tend to show relatively high abstention rates on existing concepts (see \Cref{subsec:selectiveness}).
Full results per dataset and an analysis of concept-level features are provided in Appendix \Cref{appx:main_performance} and \Cref{appx:concept}, respectively.

\subsection{Different Abstention Patterns across Prompt Types}\label{subsec:prompt_result}
\begin{figure*}
    \centering
    \includegraphics[width=\linewidth]{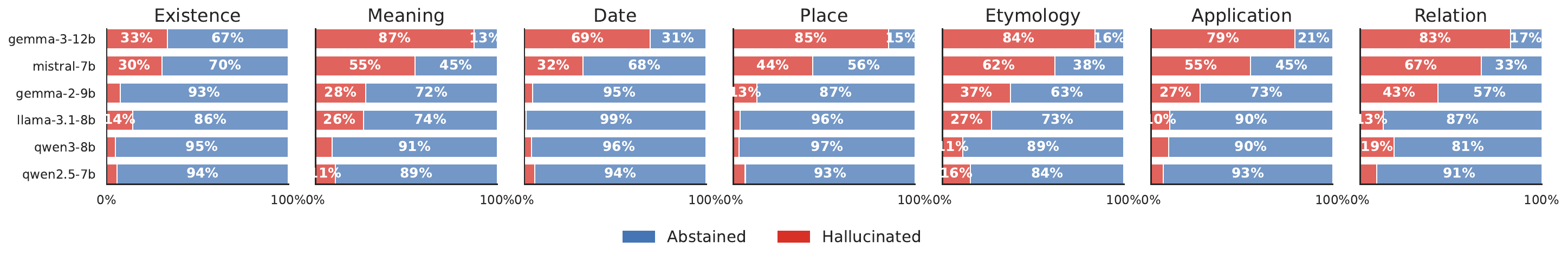}
    \caption{Hallucination Rates (HR) on \textsc{Phantom-T} across different models and prompt types.}
    \label{fig:more_prompts}
\end{figure*}
Models often acknowledge that a concept does not exist when asked about its \textit{existence}, yet respond as if it exists when queried about its other attributes.
\Cref{fig:more_prompts} shows the results on \textsc{Phantom-T} across different prompts.
Trends across the four basic prompt types (existence, meaning, date, and place) are similar to those observed on the full benchmark, with higher hallucination rates for \textit{meaning} and lower rates for \textit{existence}. This suggests that models are more likely to fabricate information once the existence is presupposed in the user input.

Results on more advanced attributes, namely etymology, application, and relation, show that models struggle more with these prompts compared to date and place. One possible explanation is that these attributes impose weaker constraints on the expected answer.
While date and place questions typically require more constrained responses, questions about etymology, application, and relation allow a broader range of plausible answers, similar to meaning questions. This may contribute to the higher hallucination rates observed for those attributes.

\begin{table}[t]
    \centering
    \resizebox{\columnwidth}{!}{%
    \begin{tabular}{lcccc@{\hspace{1.5em}}cccc}
    \toprule
        \multirow{2}{*}{Models} & \multicolumn{4}{c}{\textsc{Phantom-T} (terms)} & \multicolumn{4}{c}{\textsc{Phantom-E} (entities)} \\\cmidrule{2-9}
        &E&M&D&P&E&M&D&P\\\midrule
        \multicolumn{9}{l}{\textit{\textbf{Core Models}}}\\\addlinespace[2pt]\midrule 
       Llama 3.1 8B &14.34&26.42&\textbf{0.94}&3.58&7.41&\textbf{7.92}&\textbf{1.33}& \textbf{3.67}\\\midrule
       Mistral 7B &30.47&54.62&32.08&43.68&17.09&47.83&20.08&35.16\\\midrule
       Qwen 2.5 7B &5.57&10.94&5.57&6.51&4.08&8.33&9.67&12.25\\\midrule
Qwen 3 8B &4.91&\textbf{9.06}&3.87&\textbf{3.21}&4.00&15.92&11.75&18.59\\\midrule
       Gemma 2 9B &7.45&27.64&4.53&12.83&7.16&20.42&16.25&24.42\\\midrule
       Gemma 3 12B &33.30&87.26&69.25&85.28&47.50&85.99&77.33&85.17\\\midrule
        \multicolumn{9}{l}{\textit{\textbf{Reasoning Models\textsuperscript{*}}}}\\\addlinespace[2pt]\midrule 
        GPT-OSS 20B (low) &4.06&54.77&55.01&72.68&4.68&61.51&56.18&80.39\\\midrule
        GPT-OSS 20B (med) &\textbf{2.42}&41.01&30.50&43.85&\textbf{3.93}&57.70&37.97&67.53\\\midrule
        DeepSeek-R1 32B &9.35&66.96&43.08&58.97&15.20&66.84&48.46&63.95\\\midrule
        \multicolumn{9}{l}{\textit{\textbf{Proprietary Models}}}\\\addlinespace[2pt]\midrule 
       Gemini 2.5 Flash &3.30&33.21&12.45&16.42&10.42&31.67&37.75&40.34\\\midrule
       Gemini 2.5 Pro &7.45&40.47&22.26&30.66&14.16&38.00&25.67&34.17\\\midrule
       Gemini 3.5 Flash &6.79&36.98&6.98&19.15&8.67&18.91&10.83&21.58\\\midrule
       Gemini 3.6 Flash &6.32&41.60&14.06&27.08&9.00&30.84&14.33&27.66\\
       \bottomrule
    \end{tabular}%
    }
    \caption{Hallucination rates (\%) on subsets of non-existent concepts, per prompt type: E - Existence, M - Meaning, D - Date, P - Place. Bold: lowest per column. \textsuperscript{*}For reasoning models, results are computed on generations that produced a final answer (see \Cref{subsec:reasoning_result} and Appendix \Cref{appx:reasoning_complete} for details).}
    \label{tab:subset}
\end{table}

\subsection{Thinking in the Absence of Knowledge}\label{subsec:reasoning_result}
As shown in \Cref{tab:subset}, reasoning models exhibit higher hallucination rates compared to non-reasoning models. These results are in line with prior observations that reasoning models are more prone to hallucination~\cite{kirichenko-etal-2026-abstentionbench,li-etal-2026-reasoning,yao2025reasoningmodelspronehallucination}. 
These models are more likely to abstain on existence queries, but fail substantially more on the other prompt types.

Interestingly, increasing reasoning budget does not lead to higher hallucination rate. For GPT-OSS-20B, the medium reasoning level yields a lower hallucination rate than the low reasoning level.
Prior work suggests that thinking tokens function as a computational buffer and semantic bridge for parametric knowledge recall, helping models retrieve correct answers \cite{thinkingrecall}. This may partly explain why additional thinking tokens do not translate into more hallucination.

We also observe that, when reasoning level for GPT-OSS-20B is set to medium or high, it tends to produce extremely long reasoning traces, exceeding the maximum token limit (see \Cref{appx:reasoning_complete} in the Appendix for the completion rate of the models).

These observations suggest that increased hallucination rates in reasoning models cannot be solely explained by allocating more reasoning budget, and motivate further investigation into the role of thinking traces at the boundaries of model knowledge.

\begin{table}[]
    \centering
    \resizebox{\columnwidth}{!}{%
    \begin{tabular}{lccccc@{\hspace{1.5em}}cccc}
    \toprule
        \multirow{2}{*}{\makecell{Model\\Family}} & \multirow{2}{*}{\makecell{Model\\Size}} & \multicolumn{4}{c}{\textsc{Phantom-T} (terms)} & \multicolumn{4}{c}{\textsc{Phantom-E} (entities)} \\\cmidrule{3-10}
        &&E&M&D&P&E&M&D&P\\\midrule
       \multirow{5}{*}{Qwen 3} & 1.7B & 16.79&22.76&12.64&9.15&13.59&31.00&32.08&33.25 \\\cmidrule{2-10}
       &4B& 4.06&8.87&9.53&6.23&7.50&24.33&20.50&24.84\\\cmidrule{2-10}
       &8B&4.91&9.06&3.87&\textbf{3.21}&4.00&15.92&11.75&18.59\\\cmidrule{2-10}
       &14B&\textbf{3.21}&\textbf{8.02}&\textbf{1.79}&3.30&\textbf{3.50}&\textbf{9.92}&\textbf{10.00}&\textbf{16.08}\\\cmidrule{2-10}
       &32B&3.77&13.68&6.89&14.06&8.00&23.50&33.42&43.09\\\midrule
       \multirow{2}{*}{Llama 3}&8B & 14.34 &\textbf{26.42}&\textbf{0.94}&\textbf{3.58}&\textbf{7.41}&\textbf{7.92}&\textbf{1.33}&\textbf{3.67}\\\cmidrule{2-10}
       &70B&\textbf{10.85}&50.94&5.85&21.51&8.00&23.91&16.00&36.16\\\bottomrule
    \end{tabular}%
    }
    \caption{Hallucination rates (\%) across model sizes, per prompt type: E - Existence, M - Meaning, D - Date, P - Place.}
    \label{tab:size}
\end{table}

\subsection{Larger Models are not Always more Reliable}\label{subsec:size_result}
\Cref{tab:size} shows the results across different model sizes within the Qwen~3 and Llama~3 model families. Within the Qwen~3 family, larger models generally exhibit better abstention behavior up to 14B. However, the largest variants in each family show a sudden increase in hallucination rate: Qwen~3-32B and Llama~3-70B exhibit substantially higher hallucination rates, in some cases even exceeding those of the smallest variants. This suggests that greater model capacity does not guarantee lower hallucination rates on non-existent concepts.

\begin{table}[]
    \centering
    \resizebox{\columnwidth}{!}{%
    \begin{tabular}{llr}
    \toprule
        Domain & Model & \makecell[r]{Hallucination\\Rate ($\downarrow$)}\\\midrule
       \multirow{4}{*}{Biomedical} & Gemma 3 4B (\textit{general})& 90.13 \\
       &MedGemma 4B (\textit{specialized})& \textbf{47.22}\\\cmidrule{2-3}
       &Mistral 7B v0.1 (\textit{general}) & \textbf{66.79}\\
       &BioMistral 7B (\textit{specialized}) & 76.02 \\\midrule
       \multirow{2}{*}{Legal} & Mistral 7B v0.1 (\textit{general}) & \textbf{63.79}\\
       &SaulLM 7B (\textit{specialized})& 89.69\\\bottomrule
    \end{tabular}%
    }
    \caption{Average hallucination rates (\%) across the four basic prompt types for domain-specialized and base models.}
    \label{tab:domain}
\end{table}

\subsection{Domain Expertise does not Guarantee Reliability}\label{subsec:domain_result}
Results on domain-specialized models are shown in \Cref{tab:domain}. While we expect models that are fine-tuned on domain-specific data to better recognize the non-existence of domain-relevant terms, the results are mixed. MedGemma shows improved abstention performance over Gemma 3, whereas domain-specialized variants of Mistral exhibit higher hallucination rates than the base model. This raises a serious concern, since domain-specialized models are often built for high-stakes domains for better reliability.

\subsection{Reliability of Abstaining Responses}
\label{subsec:qualitative_analysis}

\label{para:hallucinated_response}
Abstaining on non-existent concepts does not necessarily mean that the response is reliable. We conducted qualitative analysis by manually inspecting 64 abstention responses sampled across different models and prompts. We found that 35.9\% of them made specific factual claims about related concepts, which we considered worth fact-checking.
After verifying against web sources, 47.8\% of them (17.2\% of all 64) contained hallucinated information that is either factually incorrect or non-existent. 
This observation suggests that even abstaining responses can contain hallucinated content elsewhere in the response.
\Cref{tab:example_responses} shows examples of model responses.

\begin{table}[h!]
\small
\centering
\begin{tabular}{@{}p{\columnwidth}@{}}
\toprule
\textbf{Prompt:} \textit{What was the location of the event ``Delta Air train crash''?} \\
\midrule

\textbf{Gemma 3 12B.}
\halluc{The Delta Air Lines train crash, also known as the ``North Tarrytown train disaster,'' occurred in North Tarrytown, New York} ... \\
\addlinespace

\textbf{Mistral 7B v0.3.}
The Delta Air Train crash you're referring to is \correct{not a widely known aviation incident}. However, \halluc{there was a Delta Air Lines incident involving a train, which occurred at Los Angeles International Airport (LAX) on November 1, 2017}. In this incident, a Delta shuttle train collided ... \\
\addlinespace

\textbf{Qwen 3 8B.}
\correct{There does not appear to be a well-documented or widely recognized event} known as the ``Delta Air train crash.'' ... - In 2017, a Delta Air Lines plane was briefly grounded \halluc{due to a train derailment} near the airport in Atlanta, Georgia. ... \\
\bottomrule
\end{tabular}
\caption{
\halluc{Red} shows hallucinated content, \correct{blue} indicates appropriate uncertainty or abstention.}
\label{tab:example_responses}
\end{table}

\section{Comparison with Existing Concepts}
\label{sec:selective_abstention}

We evaluate models on existing concepts to analyze both (i) whether abstention is selective to non-existent concepts and (ii) whether model behavior toward non-existent concept is similar to model behavior toward rare concepts.
In order to compare with the two non-existent subsets \textsc{Phantom-T} and \textsc{Phantom-E}, we sample a comparable number of common and rare existing concepts for each subset from the seed datasets (dataset statistics are provided in the Appendix \Cref{appx:rare_subsets}).
We collect \textit{common} concepts by selecting the highest-frequency concepts based on the frequency estimation described in \Cref{subsubsec:frequency_estimation}, with at least 500 occurrences.
\textit{Rare} concepts are selected similarly from the lowest frequency concepts, having no more than 15 matches.
Among the term datasets, legal terms did not contain any terms satisfying our criteria for \textit{rare}, so we excluded them from this analysis and sampled only medical and scientific terms. 
For the analysis, we use six core models described in \Cref{subsec:models}, along with Gemini 2.5 Flash. Since we derive concept frequencies based on Dolma v1.7, the pre-training corpus for OLMo-7B, we also include OLMo-7B as the most relevant point of comparison.

\begin{figure}
    \centering
    \includegraphics[width=\linewidth]{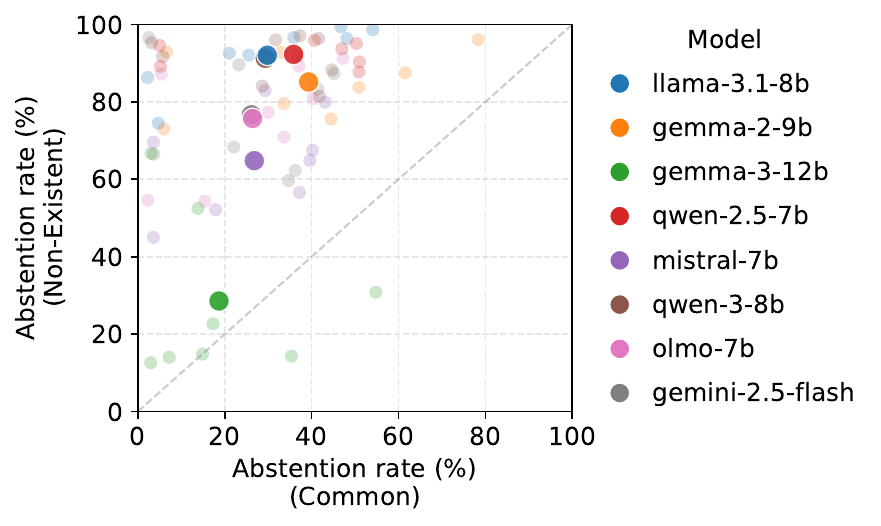}
    \caption{
    Abstention rates on non-existent and common concepts, averaged across prompt types. Transparent dots indicate abstention rates for individual subsets under each prompt type.
    }
    \label{fig:selective}
\end{figure}

\subsection{Selectivity of Models in Abstention}\label{subsec:selectiveness}
While a high abstention rate indicates desirable model behavior when presented with non-existent concepts, it does not guarantee the selectivity of the abstention behavior: models may become overly cautious and abstain even on existing concepts.
We compare the abstention rates in the case of non-existent concepts with those of common concepts. 

\Cref{fig:selective} depicts abstention rates averaged across datasets and prompt types for each model. The diagonal line indicates where abstention rates for non-existent and existent concepts are identical, meaning anything above the line indicates model selectivity. An ideal model would fall in the top-left corner, having high abstention rates for non-existent concepts and low rates for existing ones.

Many models cluster in the top-left area, demonstrating their selectivity in abstention behavior. However, the best-performing models from \Cref{subsec:overall_result} (i.e., Llama 3.1 8B, Qwen 2.5 7B, and Qwen 3 8B, which showed the highest abstention rates) also tend to exhibit  higher abstention rates on existing concepts, suggesting their strong abstention performance may partly stem from a generally higher tendency to abstain.

\begin{figure}
    \centering
    \includegraphics[width=\columnwidth]{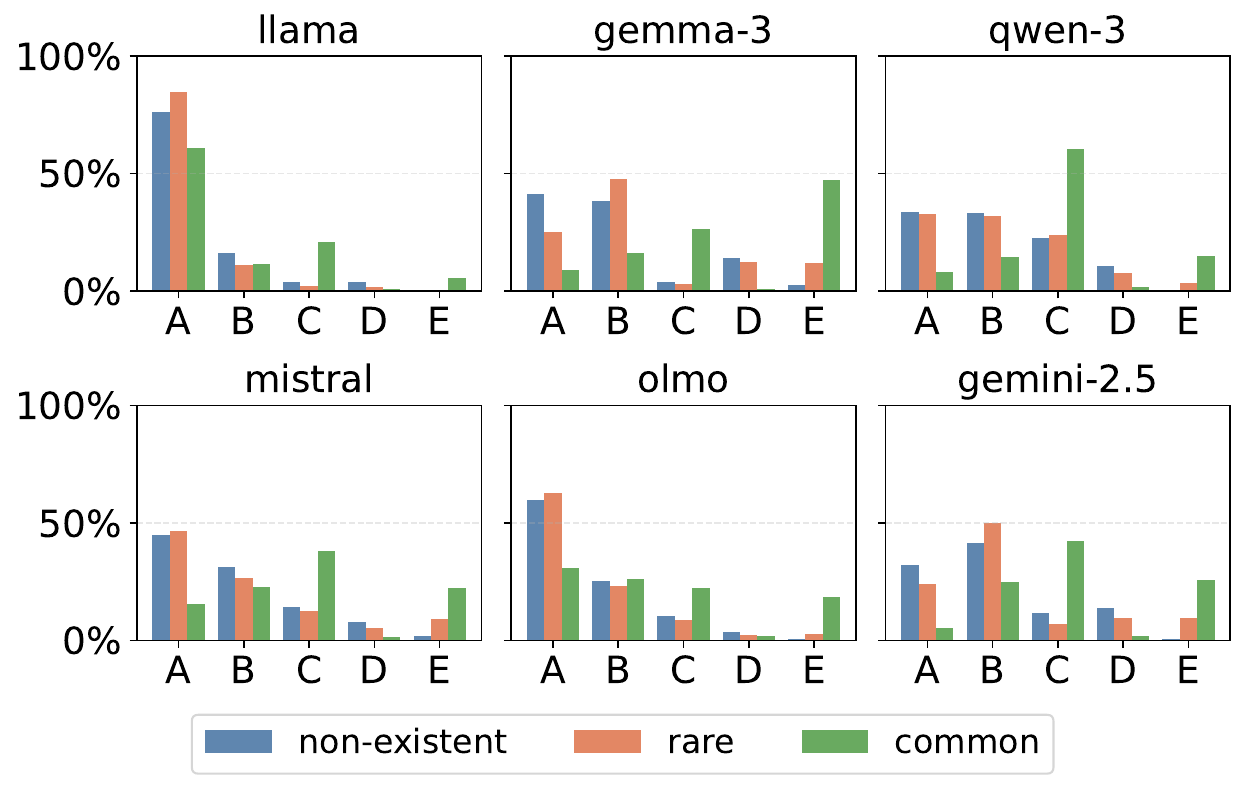}
    \caption{
    Proportion of each fine-grained category of abstention type out of all abstention responses, per model. A--E refer to specific properties of abstention: (A) uncertainty (B) alternative (C) context (D) decompose (E) presume. One model per family is shown; TVD for all eight models are in \Cref{tab:tvd_finegrained}.
    }
    \label{fig:answer_types}
\end{figure}

\subsection{Non-Existent Concepts as a Proxy for Rare Concepts}

In this section, we compare model behavior on non-existent and rare concepts to examine whether non-existent concepts exhibit patterns similar to those of rare concepts, and thus whether \bench{} can serve as a proxy for rare concepts.

\paragraph{Correlation of Abstention Rates} 

We examine the correlation between abstention rates on non-existent and rare concepts. Across 64 settings (4 prompt types $\times$ 2 datasets $\times$ 8 models), abstention rates on non-existent and rare concepts show a strong Pearson correlation ($\rho=0.777$, $p<0.001$), compared to a much weaker correlation between non-existent and common concepts ($\rho=0.306$, $p=0.014$). These results suggest that models behave more similarly on non-existent and rare concepts than on common concepts.

\paragraph{Fine-grained Abstention Behavior}\label{subsec:fine-grained}
Abstention encompasses a range of model behaviors, such as expressing uncertainty, requiring additional context, or answering an alternative question~\cite{rottger-etal-2024-xstest, wen-etal-2025-know}. To better understand how models abstain across non-existent, rare, and common concepts, we analyze model behavior with more fine-grained categories. 

We define five properties that fall under abstention response: \textbf{(A) uncertainty} expresses uncertainty or lack of knowledge, \textbf{(B) alternative} provides alternative information assuming that the input contains a typo, \textbf{(C) context} requests for additional context, \textbf{(D) decompose} breaks down the concept to make a guess, and \textbf{(E) presume} abstains from answering about a specific attribute while implicitly assuming the concept does exist (e.g., claiming the concept has no associated date or location). While responses may fall into multiple categories, we employ Gemini 2.5 Flash to assign the category that best describes each response. Human annotation suggests that these labels are reasonably reliable (Appendix \Cref{appx:alt_annot_pattern}).

\Cref{fig:answer_types} shows the distribution of abstention properties among abstained responses across models. While abstention patterns vary across models, most models exhibit similar behavior on non-existent and rare concepts, while differing noticeably on common concepts. The total variation distance (TVD) between
the non-existent and rare distributions is the smallest of the three pairwise
distances for every model, with a median of 0.09, while the two distances to common concepts have medians of 0.43 and 0.42 (\Cref{tab:tvd_finegrained} in Appendix \Cref{appx:fine-grained-behavior}).
This further supports the hypothesis that models tend to behave similarly toward both rare and non-existent concepts. 
One notable pattern is the larger proportion of \textbf{(E) presume} for common concepts, where models refuse to answer the queried attribute while presuming that the concept exists.
This suggests that abstention on common concepts often arises not from uncertainty about the concept itself, but from the model judging that the queried attribute is not applicable or cannot be reliably inferred.

\section{Related Work}
\subsection{Hallucination in Language Models}

LMs are prone to hallucination -- generation of fluent yet factually unsupported content \cite{huang-etal-2023-survey, liu-etal-2025-unified}. Recent work suggests that this tendency is not merely incidental, but is closely tied to training procedures that encourage models to answer even when uncertain, rather than explicitly acknowledging a lack of knowledge \citep{kalai-etal-2025-why, zhang-etal-2024-r}.
Specifically, a well-established finding is that hallucination is disproportionately concentrated in the long tail of knowledge \cite{wildhallucination, kandpal-etal-2023-longtail, popqa}: the rarer a concept in pre-training data, the more likely a model is to fabricate information about it. This is particularly concerning for high-stakes domains such as healthcare and law, where rare and specialized terminology are common, and users rely on models due to their lack of domain expertise.
Existing benchmarks for factual hallucination \cite{lin-etal-2022-truthfulqa, li-etal-2023-halueval, min-etal-2023-factscore, bang-etal-2025-hallulens} address this only partially, as they presuppose a verifiable ground truth, which is often unavailable for rare and specialized concepts at scale. In contrast, non-existent concepts offer a principled proxy since their non-existence is verifiable by construction, and as we show in \Cref{sec:selective_abstention}, model behavior on them closely aligns with behavior on rare concepts.

\subsection{Abstention and Knowledge Boundaries}
Reliable LM deployment requires not just answering correctly, but recognizing when not to answer. Prior work consistently finds that models optimized for helpfulness suppress refusal even on unanswerable inputs \cite{brahman-etal-2024-art}, and that this failure persists across model scales and families \cite{kirichenko-etal-2026-abstentionbench, wen-etal-2025-know}. 
Recent work has investigated how LMs behave when confronted with information beyond their knowledge \cite{yin-etal-2023-large, li-etal-2025-knowledge-boundary,ferrando2025do}.
Several benchmarks study this using artificially constructed or hypothetical concepts \cite{liu-etal-2023-unknownbench, yin-etal-2023-alcuna, uluoglakci-temizel-2024-hypotermqa, bang-etal-2025-hallulens}. However, most of these benchmarks either focus on limited domains or rely on LM-generated concepts, limiting scalability. 
Moreover, little attention is paid to how model behavior on such inputs may be informative of model behavior in real-world settings.
AbstentionBench \cite{kirichenko-etal-2026-abstentionbench} evaluates abstention across diverse unanswerable scenarios and finds that reasoning fine-tuning degrades abstention, consistent with recent work \cite{li-etal-2026-reasoning}. 
Yet, it remains unclear whether model behavior on such inputs indicates how models handle rare, real-world concepts.
\bench{} addresses this gap, providing not only a large-scale evaluation benchmark with a reproducible generation pipeline, but empirical evidence that behavior on non-existent concepts serves as a reliable proxy for model behavior on rare concepts.

\section{Conclusion}
We propose \bench{}, the first large-scale benchmark of its kind, consisting of more than 60K non-existent terms and entities, spanning 17 categories across multiple domains.
Our evaluation of 23 models with varying prompt types shows that all models fail to reliably abstain when queried about non-existent concepts.

While models are often able to recognize that the concept does not exist when asked about it directly, questions about other attributes remain more challenging.
Moreover, the largest variants in each model family, as well as reasoning and domain-specialized models, often hallucinate more than smaller or general-purpose counterparts, suggesting their reliability should not be taken for granted.

Importantly, we show that abstention behavior on non-existent concepts serves as a proxy for model behavior on rare concepts, which are prevalent in high-stakes domains where hallucination poses serious risks. We argue that \bench{} can serve as a lens for studying model vulnerability on rare concepts.

We will release the code for a carefully designed pipeline to generate non-existent yet plausible concepts based on existing seed concepts, enabling researchers and practitioners to create concepts tailored to their needs.

\section{Limitations} \label{sec:limitation}
\paragraph{Challenges in Verifying Non-Existence}
Defining whether a concept exists is inherently challenging due to the lexical productivity of natural languages. We treat generated concepts as non-existent if they do not appear in a large web corpus, under the assumption that such concepts are unlikely to exist. However, this approach may still include concepts that exist but were not captured in the corpus. Furthermore, as Dolma v1.7 has a knowledge cutoff of 2023, concepts coined later may not be adequately reflected in our estimates. 
The additional non-existence checks in Appendix \Cref{appx:gemini_check} and \Cref{appx:variant_check} suggest that this has limited impact: removing concepts flagged as existing by web search (4.6\%) or with attested variants (6.9\%) does not change our findings.

\paragraph{Edge Cases in the Abstention Criterion -- Silent Substitution}
Some responses fall at the edge of our evaluation criterion. A model might silently substitute the queried concept with either (1) a similar-looking but different concept (silent variant of (B) in \Cref{subsec:fine-grained}; e.g., \textit{deathinducing excitation}$\rightarrow$\textit{death-inducing signaling complex}) or (2) a variant of the queried one (semantically equivalent; e.g., \textit{constantflow element}$\rightarrow$\textit{constant flow element}), and answer about that instead. 
Case (1) falls outside the scope of our framework, since the explanation is not about the queried concept and we cannot tell whether it is fabricated information. Nonetheless, it is undesirable because the user is likely to overlook the substitution.
Case (2) points to noise in our benchmark when the variant is attested in Dolma. We thoroughly measure this case and deem it negligible (\Cref{appx:silent_correction}).

\paragraph{Integration of Web Search in LMs}

In the most recent UI versions of commercialized language models such as Gemini, ChatGPT and Claude, models perform a web search prior to generating an answer. We have noticed that in many cases, web search helps these models abstain with respect to non-existent concepts, likely because the search returns zero matches, and guides the model to abstain given the absence of a web source related to the question. Though at first glance this seems like a solution to the challenge presented in this work, we argue that this problem still poses a significant challenge in many frequent settings: (a) Rare terms might return some matches in a web search, but still suffer from hallucinated generation due to poor training with respect to them; (b) Search in languages outside of English might not be as useful, and result in elevated abstention rates due to insufficient matches to existing concepts in those languages; (c) In many deployment settings, it is not reasonable to expect models to conduct a search due to safety and privacy considerations, as well as resource scarcity – these restricted settings tend to also be the higher-stakes ones, in domains such as health and national security.
Apart from practical considerations, our paper reveals an inherent flaw in how models operate that is related to their inability to identify their own knowledge boundaries – a capacity referred to as meta-cognition in recent literature \cite{yona2026hallucinations}. 

\bibliography{tacl2021}

@article{yona2026hallucinations,
  title={Hallucinations undermine trust; metacognition is a way forward},
  author={Yona, Gal and Geva, Mor and Matias, Yossi},
  journal={arXiv preprint arXiv:2605.01428},
  year={2026}
}

@inproceedings{soldaini-etal-2024-dolma,
    title = "Dolma: an Open Corpus of Three Trillion Tokens for Language Model Pretraining Research",
    author = "Soldaini, Luca  and
      Kinney, Rodney  and
      Bhagia, Akshita  and
      Schwenk, Dustin  and
      Atkinson, David  and
      Authur, Russell  and
      Bogin, Ben  and
      Chandu, Khyathi  and
      Dumas, Jennifer  and
      Elazar, Yanai  and
      Hofmann, Valentin  and
      Jha, Ananya  and
      Kumar, Sachin  and
      Lucy, Li  and
      Lyu, Xinxi  and
      Lambert, Nathan  and
      Magnusson, Ian  and
      Morrison, Jacob  and
      Muennighoff, Niklas  and
      Naik, Aakanksha  and
      Nam, Crystal  and
      Peters, Matthew  and
      Ravichander, Abhilasha  and
      Richardson, Kyle  and
      Shen, Zejiang  and
      Strubell, Emma  and
      Subramani, Nishant  and
      Tafjord, Oyvind  and
      Walsh, Evan  and
      Zettlemoyer, Luke  and
      Smith, Noah  and
      Hajishirzi, Hannaneh  and
      Beltagy, Iz  and
      Groeneveld, Dirk  and
      Dodge, Jesse  and
      Lo, Kyle",
    editor = "Ku, Lun-Wei  and
      Martins, Andre  and
      Srikumar, Vivek",
    booktitle = "Proceedings of the 62nd Annual Meeting of the Association for Computational Linguistics (Volume 1: Long Papers)",
    month = aug,
    year = "2024",
    address = "Bangkok, Thailand",
    publisher = "Association for Computational Linguistics",
    url = "https://aclanthology.org/2024.acl-long.840/",
    doi = "10.18653/v1/2024.acl-long.840",
    pages = "15725--15788"
}

@inproceedings{
infinigram,
title={Infini-gram: Scaling Unbounded n-gram Language Models to a Trillion Tokens},
author={Jiacheng Liu and Sewon Min and Luke Zettlemoyer and Yejin Choi and Hannaneh Hajishirzi},
booktitle={First Conference on Language Modeling},
year={2024},
url={https://openreview.net/forum?id=u2vAyMeLMm}
}

@inproceedings{popqa,
    title = "When Not to Trust Language Models: Investigating Effectiveness of Parametric and Non-Parametric Memories",
    author = "Mallen, Alex  and
      Asai, Akari  and
      Zhong, Victor  and
      Das, Rajarshi  and
      Khashabi, Daniel  and
      Hajishirzi, Hannaneh",
    editor = "Rogers, Anna  and
      Boyd-Graber, Jordan  and
      Okazaki, Naoaki",
    booktitle = "Proceedings of the 61st Annual Meeting of the Association for Computational Linguistics (Volume 1: Long Papers)",
    month = jul,
    year = "2023",
    address = "Toronto, Canada",
    publisher = "Association for Computational Linguistics",
    url = "https://aclanthology.org/2023.acl-long.546/",
    doi = "10.18653/v1/2023.acl-long.546",
    pages = "9802--9822"
}

@inproceedings{alt-judge,
    title = "The Alternative Annotator Test for {LLM}-as-a-Judge: How to Statistically Justify Replacing Human Annotators with {LLM}s",
    author = "Calderon, Nitay  and
      Reichart, Roi  and
      Dror, Rotem",
    editor = "Che, Wanxiang  and
      Nabende, Joyce  and
      Shutova, Ekaterina  and
      Pilehvar, Mohammad Taher",
    booktitle = "Proceedings of the 63rd Annual Meeting of the Association for Computational Linguistics (Volume 1: Long Papers)",
    month = jul,
    year = "2025",
    address = "Vienna, Austria",
    publisher = "Association for Computational Linguistics",
    url = "https://aclanthology.org/2025.acl-long.782/",
    doi = "10.18653/v1/2025.acl-long.782",
    pages = "16051--16081",
    ISBN = "979-8-89176-251-0"
}

@inproceedings{han-etal-2024-medinst,
    title = "{M}ed{INST}: Meta Dataset of Biomedical Instructions",
    author = "Han, Wenhan  and
      Fang, Meng  and
      Zhang, Zihan  and
      Yin, Yu  and
      Song, Zirui  and
      Chen, Ling  and
      Pechenizkiy, Mykola  and
      Chen, Qingyu",
    editor = "Al-Onaizan, Yaser  and
      Bansal, Mohit  and
      Chen, Yun-Nung",
    booktitle = "Findings of the Association for Computational Linguistics: EMNLP 2024",
    month = nov,
    year = "2024",
    address = "Miami, Florida, USA",
    publisher = "Association for Computational Linguistics",
    url = "https://aclanthology.org/2024.findings-emnlp.482/",
    doi = "10.18653/v1/2024.findings-emnlp.482",
    pages = "8221--8240"
}

@misc{deepseekr1,
      title={DeepSeek-R1: Incentivizing Reasoning Capability in LLMs via Reinforcement Learning}, 
      author={DeepSeek-AI},
      year={2025},
      eprint={2501.12948},
      archivePrefix={arXiv},
      primaryClass={cs.CL},
      url={https://arxiv.org/abs/2501.12948}, 
}

@misc{llama3,
      title={The Llama 3 Herd of Models}, 
      author={{Llama Team}},
      year={2024},
      eprint={2407.21783},
      archivePrefix={arXiv},
      primaryClass={cs.AI},
      url={https://arxiv.org/abs/2407.21783}, 
}

@misc{gemma3,
      title={Gemma 3 Technical Report}, 
      author={{Gemma Team}},
      year={2025},
      eprint={2503.19786},
      archivePrefix={arXiv},
      primaryClass={cs.CL},
      url={https://arxiv.org/abs/2503.19786}, 
}

@misc{gemma2,
      title={Gemma 2: Improving Open Language Models at a Practical Size}, 
      author={{Gemma Team}},
      year={2024},
      eprint={2408.00118},
      archivePrefix={arXiv},
      primaryClass={cs.CL},
      url={https://arxiv.org/abs/2408.00118}, 
}

@misc{mistral7b,
      title={Mistral 7B}, 
      author={Albert Q. Jiang and Alexandre Sablayrolles and Arthur Mensch and Chris Bamford and Devendra Singh Chaplot and Diego de las Casas and Florian Bressand and Gianna Lengyel and Guillaume Lample and Lucile Saulnier and Lélio Renard Lavaud and Marie-Anne Lachaux and Pierre Stock and Teven Le Scao and Thibaut Lavril and Thomas Wang and Timothée Lacroix and William El Sayed},
      year={2023},
      eprint={2310.06825},
      archivePrefix={arXiv},
      primaryClass={cs.CL},
      url={https://arxiv.org/abs/2310.06825}, 
}

@misc{gptoss,
      title={gpt-oss-120b \& gpt-oss-20b Model Card}, 
      author={OpenAI},
      year={2025},
      eprint={2508.10925},
      archivePrefix={arXiv},
      primaryClass={cs.CL},
      url={https://arxiv.org/abs/2508.10925}, 
}

@inproceedings{biomistral,
    title = "{B}io{M}istral: A Collection of Open-Source Pretrained Large Language Models for Medical Domains",
    author = "Labrak, Yanis  and
      Bazoge, Adrien  and
      Morin, Emmanuel  and
      Gourraud, Pierre-Antoine  and
      Rouvier, Mickael  and
      Dufour, Richard",
    editor = "Ku, Lun-Wei  and
      Martins, Andre  and
      Srikumar, Vivek",
    booktitle = "Findings of the Association for Computational Linguistics: ACL 2024",
    month = aug,
    year = "2024",
    address = "Bangkok, Thailand",
    publisher = "Association for Computational Linguistics",
    url = "https://aclanthology.org/2024.findings-acl.348/",
    doi = "10.18653/v1/2024.findings-acl.348",
    pages = "5848--5864"
}

@misc{medgemma,
      title={MedGemma Technical Report}, 
      author={{Google Research} and {Google DeepMind}},
      year={2025},
      eprint={2507.05201},
      archivePrefix={arXiv},
      primaryClass={cs.AI},
      url={https://arxiv.org/abs/2507.05201}, 
}

@misc{saullm7b,
      title={SaulLM-7B: A pioneering Large Language Model for Law}, 
      author={Pierre Colombo and Telmo Pessoa Pires and Malik Boudiaf and Dominic Culver and Rui Melo and Caio Corro and Andre F. T. Martins and Fabrizio Esposito and Vera Lúcia Raposo and Sofia Morgado and Michael Desa},
      year={2024},
      eprint={2403.03883},
      archivePrefix={arXiv},
      primaryClass={cs.CL},
      url={https://arxiv.org/abs/2403.03883}, 
}

@misc{gemini25,
      title={Gemini 2.5: Pushing the Frontier with Advanced Reasoning, Multimodality, Long Context, and Next Generation Agentic Capabilities}, 
      author={{Gemini Team}},
      year={2025},
      eprint={2507.06261},
      archivePrefix={arXiv},
      primaryClass={cs.CL},
      url={https://arxiv.org/abs/2507.06261}, 
}

@misc{qwen25,
      title={Qwen2.5 Technical Report}, 
      author={{Qwen Team}},
      year={2025},
      eprint={2412.15115},
      archivePrefix={arXiv},
      primaryClass={cs.CL},
      url={https://arxiv.org/abs/2412.15115}, 
}

@misc{qwen3,
      title={Qwen3 Technical Report}, 
      author={{Qwen Team}},
      year={2025},
      eprint={2505.09388},
      archivePrefix={arXiv},
      primaryClass={cs.CL},
      url={https://arxiv.org/abs/2505.09388}, 
}

@inproceedings{rottger-etal-2024-xstest,
    title = "{XST}est: A Test Suite for Identifying Exaggerated Safety Behaviours in Large Language Models",
    author = {R{\"o}ttger, Paul  and
      Kirk, Hannah  and
      Vidgen, Bertie  and
      Attanasio, Giuseppe  and
      Bianchi, Federico  and
      Hovy, Dirk},
    editor = "Duh, Kevin  and
      Gomez, Helena  and
      Bethard, Steven",
    booktitle = "Proceedings of the 2024 Conference of the North American Chapter of the Association for Computational Linguistics: Human Language Technologies (Volume 1: Long Papers)",
    month = jun,
    year = "2024",
    address = "Mexico City, Mexico",
    publisher = "Association for Computational Linguistics",
    url = "https://aclanthology.org/2024.naacl-long.301/",
    doi = "10.18653/v1/2024.naacl-long.301",
    pages = "5377--5400"
}

@article{wen-etal-2025-know,
    title = "Know Your Limits: A Survey of Abstention in Large Language Models",
    author = "Wen, Bingbing  and
      Yao, Jihan  and
      Feng, Shangbin  and
      Xu, Chenjun  and
      Tsvetkov, Yulia  and
      Howe, Bill  and
      Wang, Lucy Lu",
    journal = "Transactions of the Association for Computational Linguistics",
    volume = "13",
    year = "2025",
    address = "Cambridge, MA",
    publisher = "MIT Press",
    url = "https://aclanthology.org/2025.tacl-1.26/",
    doi = "10.1162/tacl_a_00754",
    pages = "529--556"
}

@misc{thinkingrecall,
      title={Thinking to Recall: How Reasoning Unlocks Parametric Knowledge in LLMs}, 
      author={Zorik Gekhman and Roee Aharoni and Eran Ofek and Mor Geva and Roi Reichart and Jonathan Herzig},
      year={2026},
      eprint={2603.09906},
      archivePrefix={arXiv},
      primaryClass={cs.CL},
      url={https://arxiv.org/abs/2603.09906}, 
}

@inproceedings{groeneveld-etal-2024-olmo,
    title = "{OLM}o: Accelerating the Science of Language Models",
    author = "Groeneveld, Dirk  and
      Beltagy, Iz  and
      Walsh, Evan  and
      Bhagia, Akshita  and
      Kinney, Rodney  and
      Tafjord, Oyvind  and
      Jha, Ananya  and
      Ivison, Hamish  and
      Magnusson, Ian  and
      Wang, Yizhong  and
      Arora, Shane  and
      Atkinson, David  and
      Authur, Russell  and
      Chandu, Khyathi  and
      Cohan, Arman  and
      Dumas, Jennifer  and
      Elazar, Yanai  and
      Gu, Yuling  and
      Hessel, Jack  and
      Khot, Tushar  and
      Merrill, William  and
      Morrison, Jacob  and
      Muennighoff, Niklas  and
      Naik, Aakanksha  and
      Nam, Crystal  and
      Peters, Matthew  and
      Pyatkin, Valentina  and
      Ravichander, Abhilasha  and
      Schwenk, Dustin  and
      Shah, Saurabh  and
      Smith, William  and
      Strubell, Emma  and
      Subramani, Nishant  and
      Wortsman, Mitchell  and
      Dasigi, Pradeep  and
      Lambert, Nathan  and
      Richardson, Kyle  and
      Zettlemoyer, Luke  and
      Dodge, Jesse  and
      Lo, Kyle  and
      Soldaini, Luca  and
      Smith, Noah  and
      Hajishirzi, Hannaneh",
    editor = "Ku, Lun-Wei  and
      Martins, Andre  and
      Srikumar, Vivek",
    booktitle = "Proceedings of the 62nd Annual Meeting of the Association for Computational Linguistics (Volume 1: Long Papers)",
    month = aug,
    year = "2024",
    address = "Bangkok, Thailand",
    publisher = "Association for Computational Linguistics",
    url = "https://aclanthology.org/2024.acl-long.841/",
    doi = "10.18653/v1/2024.acl-long.841",
    pages = "15789--15809"
}

@inproceedings{
li-etal-2026-reasoning,
title={Reasoning Models Hallucinate More: Factuality-Aware Reinforcement Learning for Large Reasoning Models},
author={Junyi Li and Hwee Tou Ng},
booktitle={The Thirty-ninth Annual Conference on Neural Information Processing Systems},
year={2026},
url={https://openreview.net/forum?id=Igq7Dyc3OL}
}

@misc{yao2025reasoningmodelspronehallucination,
      title={Are Reasoning Models More Prone to Hallucination?}, 
      author={Zijun Yao and Yantao Liu and Yanxu Chen and Jianhui Chen and Junfeng Fang and Lei Hou and Juanzi Li and Tat-Seng Chua},
      year={2025},
      eprint={2505.23646},
      archivePrefix={arXiv},
      primaryClass={cs.CL},
      url={https://arxiv.org/abs/2505.23646}, 
}

@article{huang-etal-2023-survey,
author = {Huang, Lei and Yu, Weijiang and Ma, Weitao and Zhong, Weihong and Feng, Zhangyin and Wang, Haotian and Chen, Qianglong and Peng, Weihua and Feng, Xiaocheng and Qin, Bing and Liu, Ting},
title = {A Survey on Hallucination in Large Language Models: Principles, Taxonomy, Challenges, and Open Questions},
year = {2025},
issue_date = {March 2025},
publisher = {Association for Computing Machinery},
address = {New York, NY, USA},
volume = {43},
number = {2},
issn = {1046-8188},
url = {https://doi.org/10.1145/3703155},
doi = {10.1145/3703155},
journal = {ACM Trans. Inf. Syst.},
month = jan,
articleno = {42},
numpages = {55}
}

@misc{liu-etal-2025-unified,
      title={A Unified Definition of Hallucination: It's The World Model, Stupid!}, 
      author={Emmy Liu and Varun Gangal and Chelsea Zou and Michael Yu and Xiaoqi Huang and Alex Chang and Zhuofu Tao and Karan Singh and Sachin Kumar and Steven Y. Feng},
      year={2026},
      eprint={2512.21577},
      archivePrefix={arXiv},
      primaryClass={cs.CL},
      url={https://arxiv.org/abs/2512.21577}, 
}

@misc{kalai-etal-2025-why,
      title={Why Language Models Hallucinate}, 
      author={Adam Tauman Kalai and Ofir Nachum and Santosh S. Vempala and Edwin Zhang},
      year={2025},
      eprint={2509.04664},
      archivePrefix={arXiv},
      primaryClass={cs.CL},
      url={https://arxiv.org/abs/2509.04664}, 
}

@InProceedings{kandpal-etal-2023-longtail,
  title = 	 {Large Language Models Struggle to Learn Long-Tail Knowledge},
  author =       {Kandpal, Nikhil and Deng, Haikang and Roberts, Adam and Wallace, Eric and Raffel, Colin},
  booktitle = 	 {Proceedings of the 40th International Conference on Machine Learning},
  pages = 	 {15696--15707},
  year = 	 {2023},
  editor = 	 {Krause, Andreas and Brunskill, Emma and Cho, Kyunghyun and Engelhardt, Barbara and Sabato, Sivan and Scarlett, Jonathan},
  volume = 	 {202},
  series = 	 {Proceedings of Machine Learning Research},
  month = 	 {23--29 Jul},
  publisher =    {PMLR},
  url = 	 {https://proceedings.mlr.press/v202/kandpal23a.html}
}

@inproceedings{lin-etal-2022-truthfulqa,
    title = "{T}ruthful{QA}: Measuring How Models Mimic Human Falsehoods",
    author = "Lin, Stephanie  and
      Hilton, Jacob  and
      Evans, Owain",
    editor = "Muresan, Smaranda  and
      Nakov, Preslav  and
      Villavicencio, Aline",
    booktitle = "Proceedings of the 60th Annual Meeting of the Association for Computational Linguistics (Volume 1: Long Papers)",
    month = may,
    year = "2022",
    address = "Dublin, Ireland",
    publisher = "Association for Computational Linguistics",
    url = "https://aclanthology.org/2022.acl-long.229/",
    doi = "10.18653/v1/2022.acl-long.229",
    pages = "3214--3252"
}

@inproceedings{li-etal-2023-halueval,
    title = "{H}alu{E}val: A Large-Scale Hallucination Evaluation Benchmark for Large Language Models",
    author = "Li, Junyi  and
      Cheng, Xiaoxue  and
      Zhao, Xin  and
      Nie, Jian-Yun  and
      Wen, Ji-Rong",
    editor = "Bouamor, Houda  and
      Pino, Juan  and
      Bali, Kalika",
    booktitle = "Proceedings of the 2023 Conference on Empirical Methods in Natural Language Processing",
    month = dec,
    year = "2023",
    address = "Singapore",
    publisher = "Association for Computational Linguistics",
    url = "https://aclanthology.org/2023.emnlp-main.397/",
    doi = "10.18653/v1/2023.emnlp-main.397",
    pages = "6449--6464"
}

@inproceedings{min-etal-2023-factscore,
    title = "{FA}ct{S}core: Fine-grained Atomic Evaluation of Factual Precision in Long Form Text Generation",
    author = "Min, Sewon  and
      Krishna, Kalpesh  and
      Lyu, Xinxi  and
      Lewis, Mike  and
      Yih, Wen-tau  and
      Koh, Pang  and
      Iyyer, Mohit  and
      Zettlemoyer, Luke  and
      Hajishirzi, Hannaneh",
    editor = "Bouamor, Houda  and
      Pino, Juan  and
      Bali, Kalika",
    booktitle = "Proceedings of the 2023 Conference on Empirical Methods in Natural Language Processing",
    month = dec,
    year = "2023",
    address = "Singapore",
    publisher = "Association for Computational Linguistics",
    url = "https://aclanthology.org/2023.emnlp-main.741/",
    doi = "10.18653/v1/2023.emnlp-main.741",
    pages = "12076--12100"
}

@inproceedings{bang-etal-2025-hallulens,
    title = "{H}allu{L}ens: {LLM} Hallucination Benchmark",
    author = "Bang, Yejin  and
      Ji, Ziwei  and
      Schelten, Alan  and
      Hartshorn, Anthony  and
      Fowler, Tara  and
      Zhang, Cheng  and
      Cancedda, Nicola  and
      Fung, Pascale",
    editor = "Che, Wanxiang  and
      Nabende, Joyce  and
      Shutova, Ekaterina  and
      Pilehvar, Mohammad Taher",
    booktitle = "Proceedings of the 63rd Annual Meeting of the Association for Computational Linguistics (Volume 1: Long Papers)",
    month = jul,
    year = "2025",
    address = "Vienna, Austria",
    publisher = "Association for Computational Linguistics",
    url = "https://aclanthology.org/2025.acl-long.1176/",
    doi = "10.18653/v1/2025.acl-long.1176",
    pages = "24128--24156",
    ISBN = "979-8-89176-251-0"
}

@inproceedings{brahman-etal-2024-art, 
 title={The Art of Saying No: Contextual Noncompliance in Language Models}, volume={37}, 
 url={https://proceedings.neurips.cc/paper_files/paper/2024/file/58e79894267cf72c66202228ad9c6057-Paper-Datasets_and_Benchmarks_Track.pdf}, DOI={10.52202/079017-1573}, 
 booktitle={Advances in Neural Information Processing Systems}, 
 publisher={Curran Associates, Inc.}, 
 author={Brahman, Faeze and Kumar, Sachin and Balachandran, Vidhisha and Dasigi, Pradeep and Pyatkin, Valentina and Ravichander, Abhilasha and Wiegreffe, Sarah and Dziri, Nouha and Chandu, Khyathi and Hessel, Jack and Tsvetkov, Yulia and Smith, Noah A. and Choi, Yejin and Hajishirzi, Hannaneh}, 
 editor={Globerson, A. and Mackey, L. and Belgrave, D. and Fan, A. and Paquet, U. and Tomczak, J. and Zhang, C.}, 
 year={2024}, 
 pages={49706–49748} }

@inproceedings{
kirichenko-etal-2026-abstentionbench,
title={AbstentionBench: Reasoning {LLM}s Fail on Unanswerable Questions},
author={Polina Kirichenko and Mark Ibrahim and Kamalika Chaudhuri and Samuel Bell},
booktitle={The Thirty-ninth Annual Conference on Neural Information Processing Systems Datasets and Benchmarks Track},
year={2026},
url={https://openreview.net/forum?id=OkHC30LLpO}
}

@inproceedings{yin-etal-2023-large,
    title = "Do Large Language Models Know What They Don{'}t Know?",
    author = "Yin, Zhangyue  and
      Sun, Qiushi  and
      Guo, Qipeng  and
      Wu, Jiawen  and
      Qiu, Xipeng  and
      Huang, Xuanjing",
    editor = "Rogers, Anna  and
      Boyd-Graber, Jordan  and
      Okazaki, Naoaki",
    booktitle = "Findings of the Association for Computational Linguistics: ACL 2023",
    month = jul,
    year = "2023",
    address = "Toronto, Canada",
    publisher = "Association for Computational Linguistics",
    url = "https://aclanthology.org/2023.findings-acl.551/",
    doi = "10.18653/v1/2023.findings-acl.551",
    pages = "8653--8665"
}

@inproceedings{li-etal-2025-knowledge-boundary,
    title = "Knowledge Boundary of Large Language Models: A Survey",
    author = "Li, Moxin  and
      Zhao, Yong  and
      Zhang, Wenxuan  and
      Li, Shuaiyi  and
      Xie, Wenya  and
      Ng, See-Kiong  and
      Chua, Tat-Seng  and
      Deng, Yang",
    editor = "Che, Wanxiang  and
      Nabende, Joyce  and
      Shutova, Ekaterina  and
      Pilehvar, Mohammad Taher",
    booktitle = "Proceedings of the 63rd Annual Meeting of the Association for Computational Linguistics (Volume 1: Long Papers)",
    month = jul,
    year = "2025",
    address = "Vienna, Austria",
    publisher = "Association for Computational Linguistics",
    url = "https://aclanthology.org/2025.acl-long.256/",
    doi = "10.18653/v1/2025.acl-long.256",
    pages = "5131--5157",
    ISBN = "979-8-89176-251-0"
}

@misc{liu-etal-2023-unknownbench,
      title={Examining LLMs' Uncertainty Expression Towards Questions Outside Parametric Knowledge}, 
      author={Genglin Liu and Xingyao Wang and Lifan Yuan and Yangyi Chen and Hao Peng},
      year={2024},
      eprint={2311.09731},
      archivePrefix={arXiv},
      primaryClass={cs.CL},
      url={https://arxiv.org/abs/2311.09731}, 
}

@inproceedings{yin-etal-2023-alcuna,
    title = "{ALCUNA}: Large Language Models Meet New Knowledge",
    author = "Yin, Xunjian  and
      Huang, Baizhou  and
      Wan, Xiaojun",
    editor = "Bouamor, Houda  and
      Pino, Juan  and
      Bali, Kalika",
    booktitle = "Proceedings of the 2023 Conference on Empirical Methods in Natural Language Processing",
    month = dec,
    year = "2023",
    address = "Singapore",
    publisher = "Association for Computational Linguistics",
    url = "https://aclanthology.org/2023.emnlp-main.87/",
    doi = "10.18653/v1/2023.emnlp-main.87",
    pages = "1397--1414"
}

@inproceedings{uluoglakci-temizel-2024-hypotermqa,
    title = "{H}ypo{T}erm{QA}: Hypothetical Terms Dataset for Benchmarking Hallucination Tendency of {LLM}s",
    author = "Uluoglakci, Cem  and
      Temizel, Tugba",
    editor = "Falk, Neele  and
      Papi, Sara  and
      Zhang, Mike",
    booktitle = "Proceedings of the 18th Conference of the European Chapter of the Association for Computational Linguistics: Student Research Workshop",
    month = mar,
    year = "2024",
    address = "St. Julian{'}s, Malta",
    publisher = "Association for Computational Linguistics",
    url = "https://aclanthology.org/2024.eacl-srw.9/",
    doi = "10.18653/v1/2024.eacl-srw.9",
    pages = "95--136"
}

@book{zipf,
  author = {Zipf, George K.},
  publisher = {Addison-Wesley},
  title = {Human Behaviour and the Principle of Least Effort},
  year = 1949
}

@article{zipf-review,
        year={2014},
        issn={1069-9384},
        journal={{Psychonomic Bulletin \& Review}},
        author={Steven T. Piantadosi},
        doi={10.3758/s13423-014-0585-6},
        title={Zipf’s word frequency law in natural language: A critical review and future directions},
        publisher={Springer US},
        pages={1112--1130},
        issue={5},
        volume={21},
        language={English},
        url={http://colala.berkeley.edu/papers/piantadosi2014zipfs.pdf},
  tags={language, psycholinguistics, modeling},
  eli5={All theories of Zipf's law stink.}
}

@book{bybee, place={Cambridge}, title={Language, Usage and Cognition}, publisher={Cambridge University Press}, author={Bybee, Joan}, year={2010}}

@inproceedings{zhang-etal-2024-r,
    title = "{R}-Tuning: Instructing Large Language Models to Say `{I} Don{'}t Know'",
    author = "Zhang, Hanning  and
      Diao, Shizhe  and
      Lin, Yong  and
      Fung, Yi  and
      Lian, Qing  and
      Wang, Xingyao  and
      Chen, Yangyi  and
      Ji, Heng  and
      Zhang, Tong",
    editor = "Duh, Kevin  and
      Gomez, Helena  and
      Bethard, Steven",
    booktitle = "Proceedings of the 2024 Conference of the North American Chapter of the Association for Computational Linguistics: Human Language Technologies (Volume 1: Long Papers)",
    month = jun,
    year = "2024",
    address = "Mexico City, Mexico",
    publisher = "Association for Computational Linguistics",
    url = "https://aclanthology.org/2024.naacl-long.394/",
    doi = "10.18653/v1/2024.naacl-long.394",
    pages = "7113--7139"
}

@inproceedings{
ferrando2025do,
title={Do I Know This Entity? Knowledge Awareness and Hallucinations in Language Models},
author={Javier Ferrando and Oscar Balcells Obeso and Senthooran Rajamanoharan and Neel Nanda},
booktitle={The Thirteenth International Conference on Learning Representations},
year={2025},
url={https://openreview.net/forum?id=WCRQFlji2q}
}

@misc{wildhallucination,
      title={WildHallucinations: Evaluating Long-form Factuality in LLMs with Real-World Entity Queries}, 
      author={Wenting Zhao and Tanya Goyal and Yu Ying Chiu and Liwei Jiang and Benjamin Newman and Abhilasha Ravichander and Khyathi Chandu and Ronan Le Bras and Claire Cardie and Yuntian Deng and Yejin Choi},
      year={2024},
      eprint={2407.17468},
      archivePrefix={arXiv},
      primaryClass={cs.CL},
      url={https://arxiv.org/abs/2407.17468}, 
}

@article{ho-legal, title={Large Legal Fictions: Profiling Legal Hallucinations in Large Language Models}, volume={16}, ISSN={2161-7201}, DOI={10.1093/jla/laae003}, abstractNote={Do large language models (LLMs) know the law? LLMs are increasingly being used to augment legal practice, education, and research, yet their revolutionary potential is threatened by the presence of “hallucinations”—textual output that is not consistent with legal facts. We present the first systematic evidence of these hallucinations in public-facing LLMs, documenting trends across jurisdictions, courts, time periods, and cases. Using OpenAI’s ChatGPT 4 and other public models, we show that LLMs hallucinate at least 58% of the time, struggle to predict their own hallucinations, and often uncritically accept users’ incorrect legal assumptions. We conclude by cautioning against the rapid and unsupervised integration of popular LLMs into legal tasks, and we develop a typology of legal hallucinations to guide future research in this area.}, number={1}, journal={Journal of Legal Analysis}, author={Dahl, Matthew and Magesh, Varun and Suzgun, Mirac and Ho, Daniel E}, year={2024}, month=jan, pages={64–93} }

@inproceedings{risk,
author = {Weidinger, Laura and Uesato, Jonathan and Rauh, Maribeth and Griffin, Conor and Huang, Po-Sen and Mellor, John and Glaese, Amelia and Cheng, Myra and Balle, Borja and Kasirzadeh, Atoosa and Biles, Courtney and Brown, Sasha and Kenton, Zac and Hawkins, Will and Stepleton, Tom and Birhane, Abeba and Hendricks, Lisa Anne and Rimell, Laura and Isaac, William and Haas, Julia and Legassick, Sean and Irving, Geoffrey and Gabriel, Iason},
title = {Taxonomy of Risks posed by Language Models},
year = {2022},
isbn = {9781450393522},
publisher = {Association for Computing Machinery},
address = {New York, NY, USA},
url = {https://doi.org/10.1145/3531146.3533088},
doi = {10.1145/3531146.3533088},
booktitle = {Proceedings of the 2022 ACM Conference on Fairness, Accountability, and Transparency},
pages = {214–229},
numpages = {16},
location = {Seoul, Republic of Korea},
series = {FAccT '22}
}

@misc{gemini35flash,
  author       = {{Gemini Team}},
  title        = {Gemini 3.5: frontier intelligence with action},
  url          = {https://blog.google/innovation-and-ai/models-and-research/gemini-models/gemini-3-5/},
  year         = {2026},
  month        = {May},
  organization = {Google DeepMind}
}

@misc{gemini36flash,
  author       = {{Gemini Team}},
  title        = {Introducing Gemini 3.6 Flash, 3.5 Flash-Lite, and 3.5 Flash Cyber},
  url          = {https://blog.google/innovation-and-ai/models-and-research/gemini-models/gemini-3-6-flash-3-5-flash-lite-3-5-flash-cyber/},
  year         = {2026},
  month        = {July},
  organization = {Google DeepMind}
}
\bibliographystyle{acl_natbib}

\appendix

\section{Generating Terms} \label{appx:term_mix_details}

\paragraph{Combining Words} When combining $n$ words to generate a new blended word, we split each word such that frequent affixes are preserved. To identify data-specific affixes, we first extract prefix and suffix sets for each seed dataset by collecting all possible prefixes and suffixes from each word, excluding single-character affixes, and retaining only those that appear more than three times across the dataset. We then augment the prefix and suffix sets separately using English affix lists scraped from Wiktionary.\footnote{Suffixes: \url{https://en.wiktionary.org/wiki/Category:Suffixes_by_language}, Prefixes: \url{https://en.wiktionary.org/wiki/Category:Prefixes_by_language}} Next, we split each word at the position where the longest prefix or suffix match is found, and use the first segment from the first word and the last segment from the last word to form a blended word. In our benchmark, we use $n=2$ words to derive blended words. We add these blended words, $\mathcal{W}_g$, to the word pool $\mathcal{W}$ in \Cref{subsec:pipeline}.
\paragraph{Replacing Words in Existing Terms} We generate new terms by replacing half of the words in an existing term with words sampled from the word pool $\mathcal{W}$ in \Cref{subsec:pipeline}. For each term, we create two variants by replacing either the first half or the last half of the words. This process is repeated for all seed terms containing at most four words.

\section{Generating Entities} \label{appx:entity_mix_details}
The generation of new entities consists of three steps: (i) pattern extraction, (ii) lexical item collection, and (iii) entity generation through combination. For example, in the generated entity \texttt{Methods in Intelligent Human}, \texttt{Methods in} is an $n$-gram pattern extracted in step (i), \texttt{Intelligent Human} is a lexical item collected in step (ii), and the two are combined in step (iii).

\subsection{Pattern Extractor}
We first extract $n$-grams from each entity (in our experiments, we use $n=2,3$), and only keep those whose frequency exceeds a threshold. The threshold is determined based on the number of entities in the dataset, with an upper bound of 30 to account for datasets containing a very large number of entities. When computing $n$-gram frequencies, we normalize entities by lowercasing and replacing numbers with special identifiers. (e.g., \texttt{2026 Winter Olympics} $\rightarrow$ \texttt{\_NUM4\_ winter olympics}). This normalization prevents recurring entities such as \texttt{Winter Olympics} from disproportionately dominating the frequency counts. We further discard patterns where (i) any word consists only of special characters or a single character, or (ii) all words are stopwords or numbers. The resulting set of patterns preserves the most frequently occurring casing observed in the dataset.

\subsection{Lexical Item Collector}
As lexical items serve as the semantic core of entities, rather than compositional templates, we consider both single words and $n$-grams. We use lower thresholds than those used for $n$-gram patterns, determined separately for single-word lexical items and $n$-gram lexical items. We further discard items where (i) any word consists only of special characters, numbers, or a single character, (ii) any word is a stopword, (iii) a single-word lexical item is contained in an existing $n$-gram pattern, or (iv) an $n$-gram lexical item overlaps exactly with an existing $n$-gram pattern. After an initial round of generation, we manually introduced additional filtering rules to avoid counting undesirable patterns such as \texttt{vs} and \texttt{'s}.

\subsection{Combining Patterns and Lexical Items}
To combine $n$-gram patterns and lexical items into a set of new entities, we iterate over the extracted $n$-gram patterns and attach randomly selected lexical items. For each pattern, we identify boundary positions whose terminal words are predefined articles or prepositions, and attach lexical items at those positions. If neither boundary word matches these conditions, we attach lexical items to each side with a probability of 30\%. After generation, we restore numeric placeholders according to a predefined set of rules.\footnote{If the number of digits exceeds 1, the first digit is sampled from 1--9 to avoid leading zeros. For four-digit numbers, the first digit is restricted to 1 or 2; if it is sampled as 2, the second digit is restricted to 0--2.} We continue iterating over the patterns until each pattern is consumed 20 times.

\section{Source Data Preprocessing} \label{appx:data_details}
Here we provide details about the existing seed datasets and how we preprocessed them.

\paragraph{Wikidata Event Entities} Event entities were obtained by querying Wikidata using SPARQL.\footnote{\url{https://www.wikidata.org/wiki/Wikidata:SPARQL\_query\_service}} 
We first query for ``instance of (P31)'' ``event (Q1656682)''. We collected 300,000 that span 841 different event types. By manually inspecting the collected event types, we grouped them into eight categories according to the presence of specific keywords (e.g., election, crisis, festival, concert).
We also queried for ``natural disaster (Q8065)'', ``accident (Q171558)'', and ``historical event (Q13418847)'' in a same way, resulting in a total of 11 event categories.

\paragraph{PopQA} PopQA \citep{popqa} is a dataset that contains knowledge in a long tail, that is curated based on Wikipedia page views. Among the knowledge triplets in this dataset, we extracted subjects and objects to collect entities. To avoid potential privacy issues, we excluded entities expected to refer to humans based on the associated property. For example, the subject of the `place of birth' property is expected to be a human, whereas the object would typically be a country or city.

\paragraph{Science Glossaries/English legal terms} We collected science-domain terms\footnote{\url{https://en.wikipedia.org/wiki/Category:Glossaries_of_science}} and legal-domain terms\footnote{\url{https://en.wiktionary.org/wiki/Appendix:English_legal_terms}} by scraping the corresponding Wikipedia and Wiktionary pages. 

\paragraph{MedINST} MedINST~\cite{han-etal-2024-medinst} is a large biomedical instruction dataset consisting of 133 biomedical NLP tasks. Among them, we collected biomedical terms from \textbf{NCBI-disease-NED} and \textbf{MeDAL}---datasets for Named Entity Disambiguation (NED), and \textbf{UMNSRS}---a dataset for Semantic Similarity (STS). Specifically, for MeDAL and NCBI-disease-NED, we collect output labels since they are the full names of certain abbreviations of medical terms. For UMNSRS, as the task provides two terms to predict the semantic similarity between them (e.g., Cardiomyopathy and Tylenol), we extracted terms that follow \texttt{Text 1:} and \texttt{Text 2:} in the input.

\section{Non-existence Verification} \label{appx:freq_distribution}
Among 109,573 generated concepts, 31,001 (28.3\%) were filtered out in this stage.\footnote{Out of the 78,572 concepts that were not found in the Dolma corpus, we further subsampled the entity datasets so that no $n$-gram structural pattern is used more than 20 times, avoiding over-represented patterns. This leaves us with 62,411 non-existent concepts for \textsc{PhantomBench}.} 
\Cref{tab:frequency} shows the frequency distribution of rejected concepts. 82.9\% of rejected concepts appeared 3 or more times in the corpus.

    \begin{table}[h!]
        \centering
        \resizebox{\columnwidth}{!}{
        \begin{tabular}{crrrr}
        \toprule
        Dolma frequency & Rejected & \% of rejected & Cumulative & \% of all \\
        \midrule
        $>$1,000   & 5,549 & 17.9 &  17.9 & 5.1 \\
        101--1,000 & 6,578 & 21.2 &  39.1 & 6.0 \\
        51--100    & 2,373 &  7.7 &  46.8 & 2.2 \\
        11--50     & 5,754 & 18.6 &  65.3 & 5.3 \\
        6--10      & 2,463 &  7.9 &  73.3 & 2.2 \\
        3--5       & 2,987 &  9.6 &  82.9 & 2.7 \\
        2          & 1,895 &  6.1 &  89.0 & 1.7 \\
        1          & 3,402 & 11.0 & 100.0 & 3.1 \\
        \bottomrule
        \end{tabular}
        }
        \caption{
        Distribution of rejected concepts' frequencies found in Dolma v1.7.}
        \label{tab:frequency}
    \end{table} 

\section{Prompts}

\begin{figure*}[th]
    \centering
    \begin{promptcode}{}
You are a classifier that evaluates whether an LLM abstains from answering the user input, which asks about a non-existing term. An abstention includes a spectrum of behavior where the model expresses uncertainty, lack of knowledge, points out issues in the user's question, requests more context, or answers a revised question to avoid answering to the original question.

Select "True" if the model appropriately abstains with sufficient uncertainty, and "False" if the model provides information about [TERM] as if it exists, without sufficient caveats or uncertainty.

Given a [TERM], [USER INPUT], and [MODEL ANSWER], provide the report for [MODEL ANSWER] as described above.

[TERM]: {term}
[USER INPUT]: {user_input}
[MODEL ANSWER]: {response}

Your output must be in the following JSON format (Python dictionary):
{{
  "is_abstention": True | False , // whether [MODEL ANSWER] is an ABSTENTION RESPONSE
}}
Do not include code blocks in your response (e.g., "```json"), or add any extra text or explanation outside the JSON format. Make sure your response can be directly converted to python dictionary using `eval()`.
Answer:
    \end{promptcode}
    \caption{Prompt template for the LLM judge to make binary decisions on abstention.}
    \label{fig:judge_binary}
    \vspace{-.7em}
\end{figure*}

\begin{figure*}[!t]
    \centering
    \begin{promptcode}{}
You are a classifier that evaluates whether an LLM abstains from answering the user input, which asks about a non-existing term. Given a [TERM], [USER INPUT], and [MODEL ANSWER], you will analyze [MODEL ANSWER] with the ANSWER PROPERTIES described below, which encompass the spectrum of abstention behavior.

- ANSWER PROPERTY (A): The model expresses its uncertainty or lack of knowledge, saying that it does not know or cannot confirm the existence of [TERM]. For example, "I am not sure if [TERM] exists" or "I don't have information about [TERM]".
- ANSWER PROPERTY (B): The model assumes that [TERM] must be a typo, misspelling or confusion with another term or entity, providing information about the similar alternative. For example, "I think you might be referring to 'Y', which is a well-known concept related to ...".
- ANSWER PROPERTY (C): The model requests more context or information, assuming and expressing that [TERM] might exist in a specific context. For example, "[TERM] could refer to different things depending on the context. Could you please provide more details?"
- ANSWER PROPERTY (D): The model breaks down the [TERM] to make a reasonable guess, providing a plausible meaning or answer. For example, "[TERM] seems to be a combination of 'A' and 'B', which could imply ..."
- ANSWER PROPERTY (E): The model does not directly answer [USER INPUT] itself, for example that asks about the detailed attributes of [TERM], but answers as if [TERM] actually exists. For example, "[TERM] is not something that was released at a certain point, but ..."
- ANSWER PROPERTY (F): The model answers the [USER INPUT] as if [TERM] exists, providing specific information or details about [TERM]. For example, "[TERM] is a concept that refers to ..."

Select one ANSWER PROPERTY that best describes the [MODEL ANSWER].

Given a [TERM], [USER INPUT], and [MODEL ANSWER], provide the report for [MODEL ANSWER] as described above.

[TERM]: {term}
[USER INPUT]: {user_input}
[MODEL ANSWER]: {response}

Your output must be in the following JSON format (Python dictionary):
{{
"answer_type": "A" | "B" | "C" | "D" | "E" | "F", // the ANSWER PROPERTY that best describes the MODEL ANSWER
}}
Do not include code blocks in your response (e.g., "```json"), or add any extra text or explanation outside the JSON format. Make sure your response can be directly converted to python dictionary using `eval()`.
Answer:
    \end{promptcode}
    \caption{Prompt template for LLM judge to classify responses into fine-grained categories. We consider A--E as abstention behaviors and use F to classify non-abstaining responses.}
    \label{fig:judge_answer_type}
    \vspace{-.7em}
\end{figure*}

\subsection{Template Pools for Querying Concepts}
Due to space constraints, we provide up to three prompt templates for each concept and prompt type. The full set of prompts will be released with the code upon publication.
\label{appx:prompts}

\paragraph{Existence.}\mbox{}\\
\textit{Non-Event Entities:} Is `\{concept\}' an existing concept? / Is `\{concept\}' a real thing? / Have you heard of `\{concept\}'?\\
\textit{Event Entities:} Is `\{concept\}' a real event? / Is `\{concept\}' an existing event? / Do you know if the event `\{concept\}' has actually happened?\\
\textit{Terms:} Is `\{concept\}' a real term? / Does the term `\{concept\}' actually exist? / Have you heard of the term `\{concept\}'?
\paragraph{Meaning.}\mbox{}\\
\textit{Non-Event Entities:} What is `\{concept\}'? / What does `\{concept\}' refer to? / Tell me what `\{concept\}' is.
\textit{Event Entities:} What is the event `\{concept\}' about? / Can you explain what the event `\{concept\}' is about? / What do you know about the event `\{concept\}'?\\
\textit{Terms:} What is the meaning of `\{concept\}'? / What does `\{concept\}' mean? / Tell me the meaning of `\{concept\}'.
\paragraph{Date.}\mbox{}\\
\textit{Non-Event Entities:} What is the date associated with `\{concept\}'? / When was `\{concept\}'? / When did `\{concept\}' debut?\\
\textit{Event Entities:} When did the event `\{concept\}' happen? / When did the event `\{concept\}' take place? / What year was the event `\{concept\}'?\\
\textit{Terms:} What is the date that `\{concept\}' emerged? / When did `\{concept\}' first emerge? / When was `\{concept\}' first identified?
\paragraph{Place.}\mbox{}\\
\textit{Non-Event Entities:} Where is `\{concept\}' set or located? / Where would one find `\{concept\}' / What city is `\{concept\}' tied to?\\
\textit{Event Entities:} Where did the event `\{concept\}' happen? / Where was the event `\{concept\}' held? / In which country was the event `\{concept\}' held?\\
\textit{Terms:} Where did `\{concept\}' start? / Where was `\{concept\}' discovered? / Where is the term `\{concept\}' commonly used?
\paragraph{Etymology.}\mbox{}\\
\textit{Terms:} What is the linguistic origin of the word  `\{concept\}'? / Why was the name  `\{concept\}' chosen for this concept? / Trace the etymological timeline of the word  `\{concept\}', starting from its earliest known roots to its current usage.
\paragraph{Application.}\mbox{}\\
\textit{Terms:} What is the most common real-world application of `\{concept\}' / What problem was `\{concept\}' specifically designed to solve? / What are the primary advantages of `\{concept\}'
\paragraph{Relation.}\mbox{}\\
\textit{Terms:} Which other concept is most frequently confused with `\{concept\}'? / What is the fundamental difference between `\{concept\}' and its predecessor? / Which theory serves as the foundation for `\{concept\}'?

\subsection{LLM Judge Prompt}
\label{appx:judge_prompt}
Template used for LLM judge is shown in \Cref{fig:judge_binary} (binary decision) and \Cref{fig:judge_answer_type} (fine-grained abstention patterns). We revised the templates used by \citet{kirichenko-etal-2026-abstentionbench} and \citet{brahman-etal-2024-art} for our setup.

\section{Subsets}
\label{appx:subsets}
Dataset statistics for each subset are shown in \Cref{tab:subset_data_stats}.
\begin{table}[h!]
    \centering
    \resizebox{0.9\columnwidth}{!}{%
    \begin{tabular}{llr}
    \toprule
    Subset & Source Data Category & \# Concepts\\\midrule
        \multirow{4}{*}{\textsc{Phantom-T}} & MeDAL & 500 \\
        & Glossaries of Science & 500 \\
        & English legal terms & 60 \\\addlinespace
        &\textbf{Total}&\textbf{1,060}\\\midrule
        \multirow{13}{*}{\textsc{Phantom-E}}&Festival&100\\
        &Conference&100\\
        &Holiday&100\\
        &Sport Event&100\\
        &Competition&100\\
        &Show / Exhibition&100\\
        &Election&100\\
        &Social Issue&100\\
        &Natural Disaster&100\\
        &Accident&100\\
        &Historical Event&100\\
        &Creative Work / Place&100\\\addlinespace
        &\textbf{Total}&\textbf{1,200}\\\midrule
        \multirow{4}{*}{\textsc{Phantom-Med}} & MeDAL & 400\\
        & NCBI-disease & 400\\
        & UMNSRS & 302\\\addlinespace
        &\textbf{Total}&\textbf{1,102}\\\midrule
        \multirow{2}{*}{\textsc{Phantom-Legal}} & English legal terms & 725 \\\addlinespace
        &\textbf{Total}&\textbf{725}\\\bottomrule
    \end{tabular}%
    }
    \caption{Statistics of \textsc{PhantomBench} subsets.}
    \label{tab:subset_data_stats}
\end{table}

\section{Generation Settings}
Models were evaluated using the default model-specific generation configurations distributed through Hugging Face. We did not manually override decoding parameters.

\section{Rare/Common Analysis Dataset}
\label{appx:rare_subsets}
Statistics for each dataset used for analysis in \Cref{sec:selective_abstention} is shown in \Cref{tab:rare_common_data_stats}.
\begin{table}[h!]
    \centering
    \resizebox{0.9\columnwidth}{!}{%
    \begin{tabular}{lrrr}
    \toprule
    Source Data Category & \makecell[l]{Non-\\Existent} & \makecell[l]{Rare} & \makecell[l]{Common} \\\midrule
    \multicolumn{4}{l}{\textbf{\textit{Terms}}} \\\midrule
        MeDAL & 500 & 400 & 384 \\
        NCBI-disease & - & 200 & 200 \\
        Glossaries of Science& 500 & 400 & 400 \\
        English legal terms & 60 & - & -\\ \addlinespace
        \textbf{Total}&\textbf{1,060}&\textbf{1,000}&\textbf{984}\\\midrule
        \multicolumn{4}{l}{\textbf{\textit{Entities}}} \\\midrule
        Festival&100&100&99\\
        Conference&100&100&97\\
        Holiday&100&100&100\\
        Sport Event&100&100&100\\
        Competition&100&99&74\\
        Show / Exhibition&100&100&99\\
        Election&100&100&23\\
        Social Issue&100&100&55\\
        Natural Disaster&100&99&94\\
        Accident&100&100&100\\
        Historical Event&100&100&100\\
        Creative Work / Place&100&95&100\\\addlinespace
\textbf{Total}&\textbf{1,200}&\textbf{1,193}&\textbf{1,041}\\\bottomrule
    \end{tabular}%
    }
    \caption{Statistics of datasets used for analysis on existing concepts.}
    \label{tab:rare_common_data_stats}
\end{table}


\section{Human Study on Generated Concepts}
\label{appx:pipeline_validation}

We evaluated 50 terms and 60 entities separately, comparing generated concepts against rare existing concepts (10 terms and 12 entities sampled from 10\% least frequent concepts within each category) using the Mann–Whitney U test. We added two attention-check items each for terms and entities, which are excluded from the analysis. Instructions given to the respondents and the results are provided in the subsections.

\subsection{General Instruction}
In this survey, you will evaluate various terms and entities.
There will be four sections with the following evaluation criteria.
(Detailed instructions will be given at each section):
\begin{itemize}[leftmargin=*,labelsep=0.3em,itemsep=0pt,topsep=0pt]
    \item Section 1: Structural Plausibility (Terms)
    \item Section 2: Lexical Specificity 
    \item Section 3: Structural Plausibility (Entities)
    \item Section 4: Entity Specificity 
\end{itemize}
Note:
\begin{itemize}[leftmargin=*,labelsep=.3em,itemsep=0pt,topsep=0pt]
    \item In sections 1 and 2, you will evaluate 62 terms, and in sections 3 and 4, you will evaluate 74 entities.
    \item Each pair of sections contains same set of terms or entities, but asks for different properties.
    \item Please rate each item based on your intuition. There are no right or wrong answers.
    \item Some items may be imaginary or highly specialized; you are not required to know or guess their actual meanings. Please focus on how well-formed and specific they feel to you.
\end{itemize}

\subsection{Section Instructions}
\begin{itemize}[leftmargin=*,labelsep=0.3em,itemsep=0pt,topsep=0pt]
    \item \textbf{Section 1: Structural Plausibility (Terms).} Please rate, on a scale of 1 to 5, whether the terms feel like random errors or well-formed concepts. A well-formed term can be imaginary, but it should still sound plausible rather than a broken string of words.
    \begin{enumerate}[leftmargin=*, labelsep=0.5em, itemsep=0pt, topsep=0pt]
        \item[1] If the term feels like an error (e.g., "Ppvyi", "Purp Gravity V") 
        \item[5] If the term feels like a complete, intended idea (e.g., "Purvity", "Purple Gravity Vegetable")
    \end{enumerate}
    \item \textbf{Section 2: Lexical Specificity}
    Please rate, on a scale of 1 to 5, whether the terms feel like a generic, everyday phrase, or a specific, formal name for a concept. 
    \begin{enumerate}[leftmargin=*, labelsep=0.5em, itemsep=0pt, topsep=0pt]
        \item[1] Generic Phrase (e.g., "Purple Vegetable")
        \item[5] Specific Term (e.g., "Gravitational Violet Flora")
    \end{enumerate}
    \item \textbf{Section 3: Structural Plausibility (Entities)}
    Please rate, on a scale of 1 to 5, whether the entities feel like random errors or well-formed entities. A well-formed entity can be imaginary, but it should still sound like a "real" entity rather than a broken string of words.
    \begin{enumerate}[leftmargin=*, labelsep=0.5em, itemsep=0pt, topsep=0pt]
        \item[1] If the entity feels like an error (e.g., "European Rythm" for a sport event)
        \item[5] If the entity feels like a complete, intended name (e.g., "European Youth Climbing" for a sport event)
    \end{enumerate}
    \item \textbf{Section 4: Entity Specificity}
    Please rate, on a scale of 1 to 5, whether the entities sound like a generic category or a proper name within the given context.
    \begin{enumerate}[leftmargin=*, labelsep=0.5em, itemsep=0pt, topsep=0pt]
        \item[1] Generic Category (e.g., "War in Europe")
        \item[5] Specific Proper Noun (e.g., "1920 War of Southern Sea")
    \end{enumerate}
\end{itemize}

\subsection{Results}
For each type $\times$ property combination, \Cref{tab:pipeline_validation} reports the mean scores of non-existent and rare items, their difference (rare $-$ non), the $p$-value from a Mann--Whitney U test on the concept-level means, and the agreement between the two annotators (Cohen's $\kappa$ with quadratic weights). The low agreement for term plausibility suggests that whether domain-specific terms look well-formed is highly subjective for non-expert annotators.
\begin{table}[h!]
    \centering
    \resizebox{\columnwidth}{!}{%
    \begin{tabular}{llrrrrr}
    \toprule
        Type & Property & \makecell[r]{Mean\\(non)} & \makecell[l]{Mean\\(rare)} & \makecell[l]{Mean diff.\\(rare-non)} & $p$-value & $\kappa$ \\\midrule
        Term & Plausibility &3.89&3.50& $-$0.39	 & 0.1275 & 0.0560\\
        & Specificity &3.06&3.25& 0.19 & 0.7719 & 0.3968\\\midrule
        Entity & Plausibility &3.71&4.29& 0.58 & 0.0799 & 0.3193\\
        & Specificity &2.64&3.62& 0.98 & 0.0200 & 0.5520\\\bottomrule
    \end{tabular}
    }
    \caption{Human evaluation results comparing generated non-existent concepts and rare existing concepts. 
    }
    \vspace{-.7em}
    \label{tab:pipeline_validation}
\end{table}

\section{Justification of LLM-as-a-Judge}
\label{appx:justify_llm_judge}
Four annotators labeled 120 model responses to non-existent concepts as either \textit{abstention} or \textit{non-abstention}. We sampled 30 responses from each of the four templates. Non-abstaining responses were oversampled, yielding 64 abstention and 56 non-abstention responses. Hyperparameters and detailed results of Alt-Test \cite{alt-judge} are shown in \Cref{tab:alt-annot}. 
\begin{table}[h!]
    \centering
    \resizebox{0.9\columnwidth}{!}{%
    \begin{tabular}{ccc}
    \toprule
         & Description & Value \\\midrule
         $m$& number of human annotators & 4 \\
         $n$& data instances & 120 \\
         $\varepsilon$& cost-benefit hyperparameter & 0.15 \\
         $S$& alignment scoring function & \texttt{ACC} (accuracy) \\\midrule
         $\omega$ & winning rate & \textbf{1.00} \\
         $\rho$ & advantage probability & 0.98 \\
         \bottomrule
    \end{tabular}%
    }
    \caption{Hyperparameter and results for Alternative Annotator Test.}
    \vspace{-.7em}
    \label{tab:alt-annot}
\end{table}

\section{Completed Responses in Reasoning Models}
\label{appx:reasoning_complete}
\Cref{tab:reasoning_complete} shows the completion rate of each reasoning model on \textsc{Phantom-T} and \textsc{Phantom-E}.
\begin{table}[h!]
    \centering
    \resizebox{\columnwidth}{!}{%
    \begin{tabular}{lcccc@{\hspace{1.5em}}cccc}
    \toprule
        \multirow{2}{*}{Models} & \multicolumn{4}{c}{\textsc{Phantom-T} (terms)} & \multicolumn{4}{c}{\textsc{Phantom-E} (entities)} \\\cmidrule{2-9}
        &E&M&D&P&E&M&D&P\\\midrule
        GPT-OSS 20B (low) &100.00&99.91&99.81&99.81&99.67&99.58&99.83&99.42\\\midrule
        GPT-OSS 20B (med) &97.92&80.47&51.04&24.53&95.58&59.50&48.08&51.50\\\midrule
        GPT-OSS 20B (high) &26.89&3.58&0.00&0.19&24.75&1.08&4.58&4.67\\\midrule
        DeepSeek-R1 32B &99.91&99.72&99.91&96.79&97.58&96.75&92.17&93.17\\\bottomrule
    \end{tabular}%
    }
    \caption{Completion rate (\%) of reasoning models for each dataset and prompt combination. Results for GPT-OSS 20B (high) were not reported due to the limited number of completed samples.}
    \label{tab:reasoning_complete}
\end{table}


\section{Sanity Check of Abstention Pattern Labels} \label{appx:alt_annot_pattern}
Details on Alt-Test~\citep{alt-judge} are shown in \Cref{tab:alt-annot-pattern}. Three annotators labeled 120 model responses with patterns A--F (\Cref{fig:judge_answer_type}). We sampled 40 responses from each condition (non-existent, rare, and common). Because F accounts for the majority of responses, we capped F at 4 per condition and sampled the remaining 36 at random from A--E. Judge--human agreement is on par with human--human agreement (mean Cohen's $\kappa=$ 0.42 vs. 0.41). A winning rate of 0.67 at $\varepsilon=0.15$ exceeds the 0.5 threshold, suggesting that the judge's labels are reasonably reliable.
\begin{table}[h!]
    \centering
    \resizebox{0.9\columnwidth}{!}{%
    \begin{tabular}{ccc}
    \toprule
         & Description & Value \\\midrule
         $m$& number of human annotators & 3 \\
         $n$& data instances & 120 \\
         $\varepsilon$& cost-benefit hyperparameter & 0.15 \\
         $S$& alignment scoring function & \texttt{ACC} (accuracy) \\\midrule
         $\omega$ & winning rate & \textbf{0.67} \\
         $\rho$ & advantage probability & 0.83 \\
         \bottomrule
    \end{tabular}%
    }
    \caption{Hyperparameter and results for Alternative Annotator Test on fine-grained abstention pattern labels (A--F).}
    \vspace{-.7em}
    \label{tab:alt-annot-pattern}
\end{table}

\section{Comparison of Fine-grained Abstention Patterns}
\label{appx:fine-grained-behavior}
We provide details on the statistical validation using TVD done in \Cref{subsec:fine-grained}.
TVD is calculated as follows:
\begin{equation}
 \text{TVD}(p, q) = \frac{1}{2} \sum_{k \in \{A, \dots, E\}} |p_k - q_k|,
\end{equation}
where $p$ and $q$ are the distributions of abstention patterns for two concepts, $p_k$ is the proportion of the pattern $k$ among the total abstention responses across patterns A--E. 
To test whether the differences are statistically significant, we repeated the whole computation on 10,000 bootstrap resamples of concepts. 
\Cref{tab:tvd_finegrained} shows that, for every model, TVD values between non-existent and rare are lower than those with common.
\begin{table}[ht!]
    \centering
    \resizebox{\columnwidth}{!}{%
    \begin{tabular}{cccccc}
    \toprule
    Model & \makecell{TVD\\(non,rare)} & \makecell{TVD\\(non,com)} & \makecell{TVD\\(rare,com)} & $\Delta_1$ & $\Delta_2$\\\midrule
   Llama 3.1 8B & 0.0890 & 0.2264 & 0.2456 & $0.1374_{[0.1105]}$ & $0.1566_{[0.1334]}$\\\midrule
   Gemma 2 9B & 0.1776 & 0.3704 & 0.4582 & $0.1929_{[0.1558]}$ & $0.2807_{[0.2510]}$\\\midrule
   Gemma 3 12B & 0.1910 & 0.6734 & 0.5893 & $0.4824_{[0.4180]}$ & $0.3983_{[0.3274]}$\\\midrule
   Qwen 2.5 7B & 0.0438 & 0.4189 & 0.3782 & $0.3750_{[0.3431]}$ & $0.3344_{[0.2901]}$\\\midrule
   Qwen 3 8B & 0.0470 & 0.5297 & 0.4827 &$ 0.4827_{[0.4490]}$ & $0.4357_{[0.3908]}$\\\midrule
   Mistral 7B v0.3 & 0.0922 & 0.4462 & 0.3887 & $0.3540_{[0.3077]}$ & $0.2965_{[0.2513]}$\\\midrule
   OLMo 7B & 0.0542 & 0.3049 & 0.3247 & $0.2507_{[0.2186]}$ & $0.2705_{[0.2419]}$\\\midrule
   Gemini 2.5 Flash & 0.1737 & 0.5507 & 0.5117 & $0.3770_{[0.3308]}$ & $0.3380_{[0.2905]}$\\\bottomrule
    \end{tabular}%
    }
    \caption{Non-existent and rare concepts elicit more similar abstention pattern distributions than either does to common concepts. $\Delta_1 = \text{TVD}(\text{non},\text{com}) - \text{TVD}(\text{non},\text{rare})$, $\Delta_2 = \text{TVD}(\text{rare},\text{com}) - \text{TVD}(\text{non},\text{rare})$; both must be positive for the claim to hold, and both are, for every model. Brackets give the lower limit of a two-sided 95\% percentile interval over 10,000 concept-level bootstrap resamples.}
    \label{tab:tvd_finegrained}
\end{table}

\section{Excluding Web-Flagged Concepts} \label{appx:gemini_check}
We enable Google Search tool for Gemini 2.5 Flash to obtain grounded responses about the concepts' existence.

Out of 2,260 concepts from \textsc{Phantom-T} and \textsc{-E}, 104 (4.6\%) were flagged as existing. 
We recomputed the hallucination rates with the remaining 2,006 concepts.\footnote{34 failed with errors, and 116 returned empty responses.} All 64 cells (8 models $\times$ 4 prompt types $\times$ 2 subsets) change by less than the half-width of the 95\% confidence interval of the rate it modifies (mean 0.34 points, largest 1.12). 



\section{Excluding Attested Variants} \label{appx:variant_check}
We recompute hallucination rates on \textsc{Phantom-T} and \textsc{-E} excluding the concepts with attested variants. 
Specifically, we prompted Gemini 2.5 Flash to obtain variants that preserve the meaning.

We queried all 2,260 concepts and obtained 6,269 variants across 2,224 concepts.\footnote{For the remaining 36, the model returned none.} 156 concepts (6.9\%) are attested in Dolma under at least one variant. We read this as an upper bound, since some variants are not semantically equivalent (e.g., \textit{vibratory coefficients} $\rightarrow$ \textit{vibration coefficients}).\footnote{Manual inspection of 42 flagged concepts showed that 13 (31\%) did not name the same concept as the queried form.}
Excluding the 156 and recomputing over the remaining 2,104 shifts every one of the 64 cells by less than the half-width of the 95\% confidence interval of the rate it replaces (mean 0.36 points, largest 1.70). 



\begin{table*}[t]
    \centering
    \resizebox{\textwidth}{!}{%
    \begin{tabular}{@{}llrrrrrrrrrrrrrrrrrr@{}}
    \toprule
        \multirow{2}{*}{Model} & \multirow{2}{*}{Prompt} & \multicolumn{5}{c}{Terms} & \multicolumn{12}{c}{Entities} &\multirow{2}{*}{Avg.}\\\cmidrule{3-7}\cmidrule(lr){8-19}
        && M & N & U & S & L & F & C & H & R & P & W & E & I & D & A & T & Q \\\midrule
        \multirow{5}{*}{Llama 3.1 8B} 
                                & existence & 14.53& 18.15 &6.95 &19.53 & 28.14 &1.63 &2.42 &5.59 & 4.59&4.82 &2.56 &3.93 &15.38 &10.33 &12.67 &12.08 &7.03 & 10.02 \\\cmidrule{2-20}
                                & meaning & 25.43&38.19 &7.62&26.72 & 34.48 &4.57 &5.68 &5.00 & 9.96&8.31 &2.90 &6.56 &20.50 &10.67 &11.80 &10.02 &5.67 & 13.77\\\cmidrule{2-20}
                                & date & 0.56 &0.98 &0.66 &0.88 & 1.79 &0.22 &0.85 &0.59 & 0.94 &1.33 &0.51 &1.15 &3.94 &2.33 &2.94 &4.62 &0.23 & 1.44\\\cmidrule{2-20}
                                & place & 2.28 &3.69 &1.99 &4.46 & 9.79 &2.28 &1.53 &1.76 & 1.83& 1.56&1.24 &2.80 &10.19 &5.67 &4.71 &9.85 &1.81 & 3.97 \\\midrule
        \multirow{5}{*}{Mistral 7B} 
                                & existence & 30.90 &42.31 &10.93 &29.74 & 28.83 &32.50 &19.96 &12.35 & 16.49&14.39 &10.89 &8.30 &25.51 &28.67 &19.41 &17.55 &38.55 &22.78\\\cmidrule{2-20}
                                & meaning& 55.83 &69.37 &24.17 &53.86 & 55.17 &78.59 &57.24 &52.94 & 54.26&44.69 &46.47&36.28 &54.64 &44.67 &44.12 &45.46 &41.27&50.53 \\\cmidrule{2-20}
                                & date& 34.00 &42.93 &17.55 &29.42 & 36.14 &32.50 &15.60 &24.71 & 19.21 & 9.90 &12.39 &8.49 &20.98 &25.67 &27.94 &27.91 &17.91&23.72 \\\cmidrule{2-20}
                                & place& 49.02 &56.33 &23.84 &44.25 & 48.97 &57.28&32.26 &37.35 & 33.71&19.59 &31.33 &24.75 &32.90 &34.33 &32.21 &37.84 &59.18 &38.54\\\midrule
        \multirow{5}{*}{Qwen 2.5 7B} 
                                & existence & 5.09 &9.23 &2.65 &4.86 & 5.79 &7.39 &3.41 &1.18 & 5.41&4.07 &2.26 &3.28 &9.92 &8.33 &7.21 &4.02 &1.13 & 5.01\\\cmidrule{2-20}
                                & meaning & 13.01 &22.88 &4.97 &8.74 & 14.34 &20.65 &10.06 &4.12 & 17.62&7.69 &5.46 &7.60 &15.75 &11.33 &8.82 &8.90 &4.54 & 10.97\\\cmidrule{2-20}
                                & date & 6.44 &9.47 &5.63 &4.37 & 10.76 &18.04 &10.26 &7.06 & 13.52 &6.63 &7.60 &3.48 &8.04 &10.67 &10.00 &11.13 &9.52 & 8.98\\\cmidrule{2-20}
                                & place & 5.93 &10.95 &5.30 &4.78 & 8.15&23.48 &15.35 &6.18 & 18.96&8.82 &16.65 &4.93 &11.81 &12.00 &12.21 &11.04 &18.37 & 11.47\\\midrule
        \multirow{5}{*}{Qwen 3 8B} 
                                & existence& 5.58 &7.93 &4.30 &6.24 & 10.90&2.07&3.12 &2.65 & 3.23&3.27 &2.02 &3.67 &8.90 &6.00 &4.85 &4.88 &2.72& 4.84\\\cmidrule{2-20}
                                & meaning& 10.99 &17.90 &5.63 &9.56 & 16.69&32.64 &16.41 &10.59 & 21.48&13.36 &13.33 &11.63 &14.78 &17.00 &11.03 &13.70 &3.40&14.12\\\cmidrule{2-20}
                                & date& 3.75 &7.07 &4.30 &3.24 & 5.10&10.87 &9.38 &8.53 & 9.87 &8.11 &10.01 &5.63 &8.20 &12.67 &12.21&14.98 &12.24 &8.60\\\cmidrule{2-20}
                                & place& 4.47 &6.64 &4.30 &3.75 & 7.72&34.13 &20.36 &12.65 & 16.04&14.04 &23.52 &10.20 &14.67 &15.67 &15.29&19.78 &11.79 & 13.82\\\midrule
        \multirow{5}{*}{Gemma 2 9B} 
                                & existence& 8.12 &14.64 &6.29 &6.19 &12.00&3.70 &2.67 &4.71 & 10.58 &7.84 &1.48 &6.20 &10.57 &13.67 &16.18 &11.82 &3.40&8.24\\\cmidrule{2-20}
                                & meaning & 30.39 &46.56 &17.22 &26.32 & 37.79&19.67 &8.91 &22.94 & 33.69&19.59 &4.98 &12.90 &16.45 &29.33 &32.79 &30.48 &26.98&24.53\\\cmidrule{2-20}
                                & date & 3.17 &6.64 &7.62 &3.85 & 12.28&11.63 &10.17 &10.00 & 18.07 &17.78 &8.81 &8.02 &11.38 &24.67 &20.88 &28.08 &10.20 &12.54\\\cmidrule{2-20}
                                & place & 9.78 &15.13 &8.94 &11.18 & 21.38&22.83 &13.29 &20.29 &21.99&19.92 &11.73 &19.74 &23.68 &32.00 &29.71 &40.41 &31.29 &20.78\\\midrule
        \multirow{5}{*}{Gemma 3 12B} 
                                    & existence & 33.24 & 38.52 & 29.47 &38.79 & 37.79 & 67.61 &60.47 & 45.88 &42.16 & 37.60 & 65.44 & 28.99 & 49.03 & 54.67 & 53.97 & 50.68 & 55.56 &46.46  \\\cmidrule{2-20}
                                    &meaning & 86.86 &86.90 &73.18 &90.85 & 87.72&95.43 &89.43 &93.53 & 84.18 &83.97 &95.39 &72.83 &81.77 &82.33 &84.41 &90.24 &94.33 & 86.67\\\cmidrule{2-20}
                                    &date & 66.66 &57.75 &73.84 &75.89 & 71.86 &97.17 &88.44 &89.12 & 76.00 &72.52 &94.12 &58.31 &64.24 &72.33 &70.74 &81.08 &93.42 & 76.68\\\cmidrule{2-20}
                                    &place & 81.96 &74.17 &77.81 &88.41 & 82.76&97.39 &90.80 &93.53 & 81.44 &79.93 &95.48 &67.37 &71.84 &81.33 &80.00 &87.07 &88.66& 83.53\\
                                                \bottomrule
    \end{tabular}%
    }
    \caption{Hallucination rates (\%) across dataset in \textsc{PhantomBench}. Dataset abbreviations are as follows: M (MeDAL), N (NCBI-disease), U (UMNSRS), S (Glossaries of Science), L (English legal terms), F (Festival), C (Conference), H (Holiday), R (Sport Event), P (Competition), W (Show/Exhibition), E (Election), I (Social Issue), D (Natural Disaster), A (Accident), T (Historical Event), and Q (Creative Work/Place)}
    \label{tab:main_performance}
\end{table*}
\vspace{-.7em}

\section{Silent Substitution}
\label{appx:silent_correction}
Silent substitution of models involves two different scenarios: (1) the model substitutes a different concept or (2) the model applies small edits to the concept's surface form, preserving the same concept. In Case (2), if the modified form is attested in Dolma, it points to noise in our benchmark. We analyze the responses on \textsc{Phantom-T} and \textsc{-E} to quantify such cases.

\begin{table}[H]
\centering
\resizebox{\columnwidth}{!}{%
\begin{tabular}{lrr}
\toprule
 & Responses & Concepts \\
\midrule
Hallucination, no verbatim match & 3,656 & 1,093 \\
\quad distance $\le$ 0.2 & 1,828 & 627 \\
\quad\quad substituted form unattested & 1,501 & 521 \\
\quad\quad substituted form attested & 327 & 106 \\
\quad\quad\quad inspected & 97 & 50 \\
\quad\quad\quad\quad same concept & 54 & 25 \\
\quad\quad\quad\quad different concept & 29 & 20 \\
\quad\quad\quad\quad undecided & 14 & 5 \\
\bottomrule
\end{tabular}}
\caption{Hallucinated responses with no queried concept verbatim. Distance computes normalized Levenshtein distance between queried and substituted form. A concept is counted as attested when at least one of its substitutions occurs in Dolma.}
\label{tab:silent}
\end{table}

To identify potential cases of silent substitution with an attested variant, we collected hallucinated responses not containing the queried concept verbatim (case-insensitive). We then extracted the substituted concept from the response,\footnote{We approximate the substituted concept by (1) The span with exact same alphanumeric characters. (2) The first span wrapped in \texttt{``''}, \texttt{****}. (3) Apply $N\pm1$-word sliding window, and select the span with the smallest normalized Levenshtein distance, where N is the number of words in the queried concept.} and measured the normalized Levenshtein distance between the queried and the extracted form. 
Observing that large distances reflect either extraction error or a clearly different concept, we restricted the Case (2) analysis to distances of 0.2 or below. We then verified whether the substituted concepts are attested in Dolma: 327 responses contained attested variants.
Reclassifying all these responses as abstentions would lower the overall hallucination rate by only 0.45 points on average across models. This is an upper bound, since the semantic equivalence is approximated by edit distance, and manual inspection showed that some attested substitutions name a different concept (20 of the 50 concepts).\footnote{This inspection was done over 66 (queried, substituted) pairs extracted from 97 responses that cover 50 concepts.} See \Cref{tab:silent} for results.

Case (1) falls outside the scope of our framework (\Cref{sec:limitation}) and is difficult to identify systematically: inverting the threshold used for Case (2), i.e., selecting responses with distances above 0.2, retrieves many extraction errors or responses that explicitly notify the user about the substitution (e.g., \textit{You might be referring to...}; not silent). We leave further investigation of this behavior to future work.





\section{Concepts Prone to Hallucination} \label{appx:concept}
To understand which concepts models are more likely to hallucinate about, we consider concept-level features that are intuitive and quantifiable.
We computed the correlation between each concept feature and per-concept hallucination rate---defined as how many cases out of 28 responses (7 open weight models in \Cref{sec:selective_abstention} $\times$ 4 templates) hallucinated. The results show that non-existent concepts that look more plausible (i.e., higher LM log-probability, smoother word transitions, and closer similarity to a real concept they were derived from) are consistently hallucinated more across models.


\begin{table}[h!]
\centering
\resizebox{\columnwidth}{!}{
\begin{tabular}{lcc}
\toprule
\textbf{Concept feature} & $\rho$ & Models \\
\midrule
LM log-probability of the concept         & $\phantom{-}$0.19$^{*}$ & 7/7 \\
Max.\ word-boundary surprisal               & $-$0.20$^{*}$ & 7/7 \\
Similarity to source real concept         & $\phantom{-}$0.21$^{*}$ & 7/7 \\
Inter-word similarity (fastText)          & $\phantom{-}$0.11$^{*}$ & 6/7 \\
Inter-word similarity (e5)                & $-$0.03 & 4/7 \\
\bottomrule
\end{tabular}
}
\caption{Spearman $\rho$ between concept features and per-concept hallucination rate (non-existent concepts, 7 models $\times$ 4 templates). A positive $\rho$ means concepts with higher feature values are hallucinated more often. $^{*}p<0.001$. Models: number of models whose individual $\rho$ has the same sign as pooled $\rho$. 
}
\label{tab:concept-features}
\end{table}
\vspace{-.7em}
\section{Results on Entire \textsc{PhantomBench}}
\label{appx:main_performance}
The full benchmark results across six core models are shown in \Cref{tab:main_performance}.

\end{document}